\documentclass{article}

    \usepackage[final, nonatbib]{neurips_2024}

\usepackage[utf8]{inputenc} 
\usepackage[T1]{fontenc}    
\usepackage{url}            
\usepackage{booktabs}       
\usepackage{amsfonts}       
\usepackage{nicefrac}       
\usepackage{microtype}      
\usepackage{xcolor}         

\usepackage{algorithm}
\usepackage{algorithmic}
\usepackage{amssymb}
\usepackage{amsmath}
\usepackage{multirow}
\usepackage{caption}
\usepackage{subcaption}
\usepackage{pifont}
\usepackage{makecell}
\usepackage{graphicx}
\usepackage{wrapfig}
\usepackage{cite}

\newcommand{\eg}{\textit{e.g.}}
\newcommand{\ie}{\textit{i.e.}}

\definecolor{cvprblue}{rgb}{0.21,0.49,0.74}
\usepackage[pagebackref,breaklinks,colorlinks,citecolor=cvprblue]{hyperref}

\title{Long-Tailed Out-of-Distribution Detection via Normalized Outlier Distribution Adaptation}

\author{%
  Wenjun Miao\thanks{Beihang University and Beihang Jiangxi Research Institute}\\
  Beihang University, China\\
  \texttt{miaowenjun@buaa.edu.cn}\\
  \And
  Guansong Pang\thanks{Corresponding authors: G. Pang and J. Zheng}\\
  Singapore Management University, Singapore \\
  \texttt{gspang@smu.edu.sg}\\
  \AND
  Jin Zheng\footnote[1]{} \thinspace \footnote[2]{}\\
  Beihang University, China \\ 
  \texttt{jinzheng@buaa.edu.cn}\\
  \And
  Xiao Bai\footnote[1]{}\\
  Beihang University, China\\
  \texttt{baixiao@buaa.edu.cn}\\
}

\begin{document}
\begin{sloppypar}

\maketitle

\begin{abstract}
  One key challenge in Out-of-Distribution (OOD) detection is the absence of ground-truth OOD samples during training. One principled approach to address this issue is to use samples from external datasets as outliers (\ie, pseudo OOD samples) to train OOD detectors.
  However, we find empirically that the outlier samples often present a distribution shift compared to the true OOD samples, especially in Long-Tailed Recognition (LTR) scenarios, where ID classes are heavily imbalanced, \ie, the true OOD samples exhibit very different probability distribution to the head and tailed ID classes from the outliers.
  In this work, we propose a novel approach, namely \textit{normalized outlier distribution adaptation} (AdaptOD), to tackle this distribution shift problem.
  One of its key components is \textit{dynamic outlier distribution adaptation} that effectively adapts a vanilla outlier distribution based on the outlier samples to the true OOD distribution by utilizing the OOD knowledge in the predicted OOD samples during inference.
  Further, to obtain a more reliable set of predicted OOD samples on long-tailed ID data, a novel \textit{dual-normalized energy loss} is introduced in AdaptOD, which leverages class- and sample-wise normalized energy to enforce a more balanced prediction energy on imbalanced ID samples. This helps avoid bias toward the head samples and learn a substantially better vanilla outlier distribution than existing energy losses during training. It also eliminates the need of manually tuning the sensitive margin hyperparameters in energy losses.
  Empirical results on three popular benchmarks for OOD detection in LTR show the superior performance of AdaptOD over state-of-the-art methods.
  Code is available at \renewcommand\UrlFont{\color{blue}}\url{https://github.com/mala-lab/AdaptOD}.
\end{abstract}

\section{Introduction}
Deep neural networks (DNNs) are widely known to be overconfident about what they do not know when applying them to real-world scenarios in open environments \cite{wang2020long, huang2021mos}, such as autonomous driving \cite{kendall2017uncertainties} and medical diagnosis \cite{leibig2017leveraging}. Consequently, the high-confidence predictions can misclassify out-of-distribution (OOD) samples from unknown classes as one of the known or in-distribution (ID) classes \cite{liu2020energy, yang2021semantically}. 
This issue is further amplified when ID samples exhibit a class-imbalanced distribution in Long-Tailed Recognition (LTR) scenarios. 
This is because head samples often receive similarly high-confident prediction as OOD samples, while the tail samples receive substantially lower-confident prediction, leading to an indistinguishability between OOD and head samples and the tendency of wrongly detecting tail samples as OOD samples \cite{zhu2023openmix,wang2022partial,miao2023out}.
We address the problem of \textit{long-tailed OOD detection}, aiming at ensuring LTR accuracy while rejecting unknown samples.

\begin{figure}[t!]
    \centering
    \begin{subfigure}[b]{0.24\textwidth}
        \includegraphics[width=\textwidth]{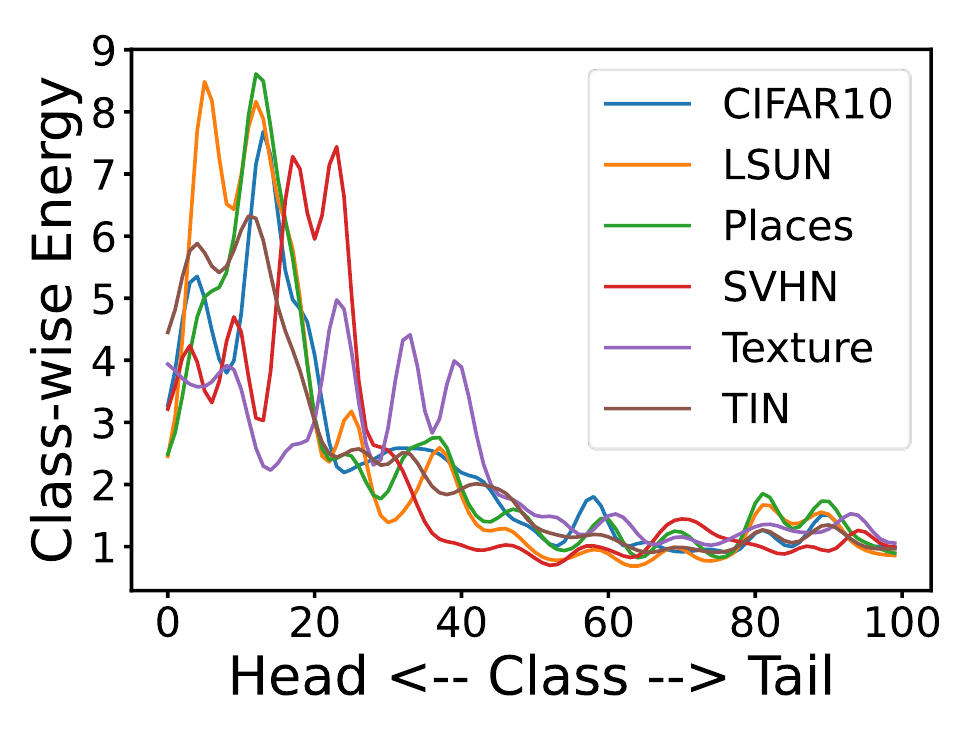}
            \caption{}
        \label{fig1:subfig1}
    \end{subfigure}
    \begin{subfigure}[b]{0.24\textwidth}
        \includegraphics[width=\textwidth]{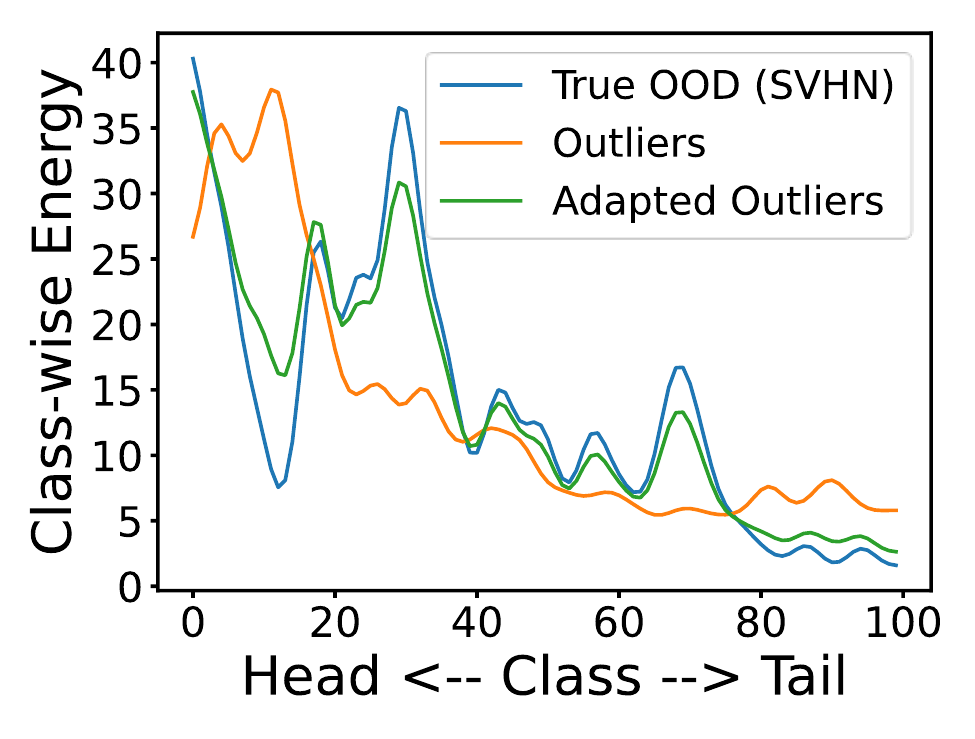}
        \caption{}
        \label{fig1:subfig2}
    \end{subfigure}
    \begin{subfigure}[b]{0.24\textwidth}
        \includegraphics[width=\textwidth]{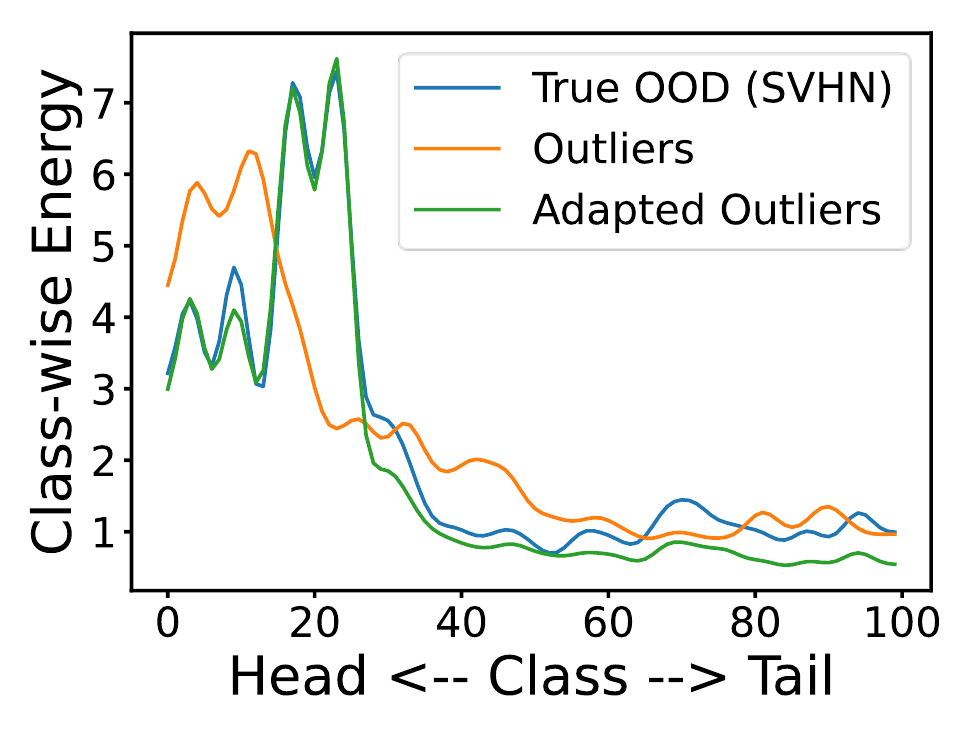}       
        \caption{}
        \label{fig1:subfig3}
    \end{subfigure}    
    \begin{subfigure}[b]{0.24\textwidth}
        \includegraphics[width=\textwidth]{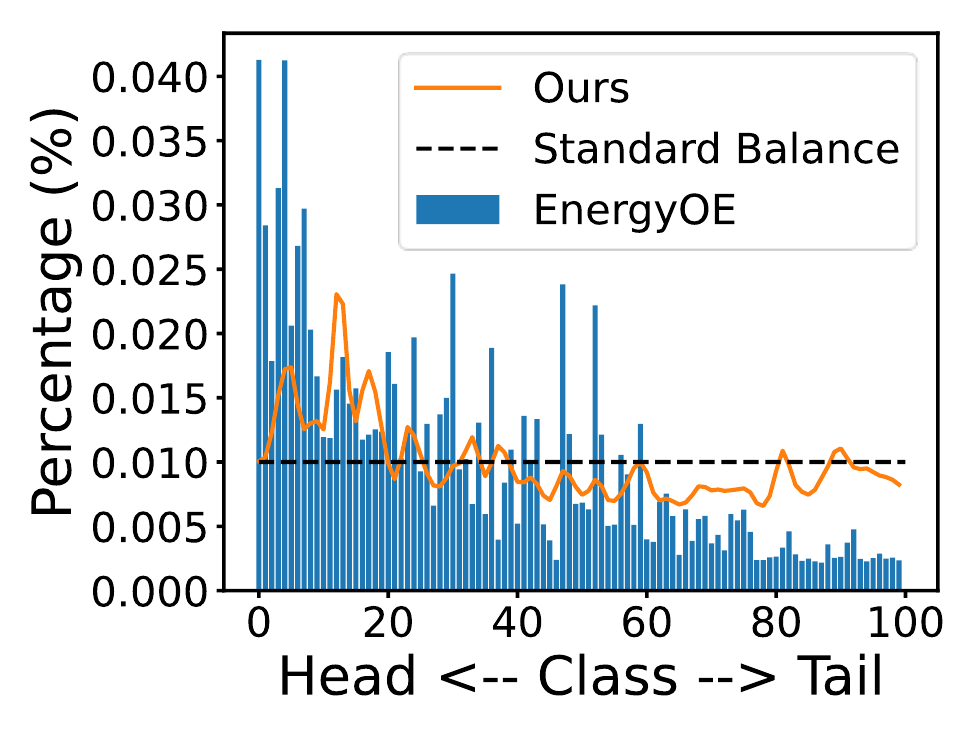}
        \caption{}
        \label{fig1:subfig4}
    \end{subfigure}
    \caption{
    \textbf{(a)} Mean energy distribution on six OOD datasets with CIFAR100-LT \cite{cao2019learning} as ID data. 
    \textbf{(b)} The results of the energy distribution of the OOD dataset SVHN \cite{netzer2011reading} using our proposed dynamic outlier distribution adaptation (DODA) and an existing energy loss EnergyOE \cite{liu2020energy}, where the outlier data is taken from TinyImages80M \cite{torralba200880}.
    \textbf{(c)} The results of using both of our proposed DODA and dual-normalized energy loss (DNE). 
    \textbf{(d)} The ratio of the energy of each ID class to the aggregated energy of all ID classes.}
    \label{fig1:main}
\end{figure}

One notorious challenge in OOD detection is the lack of ground-truth information on OOD samples, as they can be drawn from any unknown distribution. 
One popular solution to tackle this challenge is to use samples from external datasets as outliers (\ie, samples that do not overlap with ID and OOD samples, also known as pseudo OOD samples) to train OOD detectors \cite{hendrycks2018deep,liu2020energy,jiang2023detecting,choi2023balanced, miao2023out}. This approach can be implemented by fitting the prediction probability of the outlier data to a prior distribution over the ID classes \cite{jiang2023detecting} or a margin-based global energy function \cite{choi2023balanced}.
Despite showing good performance on various benchmarks, all of these methods assume that the distribution of the outliers is well aligned with that of the true OOD samples in the target data.
However, the outliers often present a distribution shift compared to the true OOD samples, especially in LTR scenarios \cite{yang2023auto,zhang2023model}, \ie, the true OOD samples exhibit very different probability distribution to the head and tailed ID classes from the outliers. Due to the bias toward head classes, the distribution shift is particularly severe w.r.t. the head samples. 
For example, as shown in Fig. \ref{fig1:subfig1}, the energy distribution of six popular OOD datasets differs significantly from each other, where CIFAR100-LT~\cite{cao2019learning} is used as the ID dataset. This implies that any of these OOD datasets used as outlier data source can largely mismatch the distribution of the true OOD data if the other five datasets are used as the true OOD data.
Such a distribution shift can largely mislead the training of detection models, leading to downgraded detection performance.

To tackle this problem, in this work, we propose a novel approach for OOD detection in LTR, namely Normalized Outlier Distribution Adaptation (\textbf{AdaptOD}). Dynamic Outlier Distribution Adaptation (\textbf{DODA}) is a key component of AdaptOD. Given a vanilla outlier distribution, DODA performs test-time adaptation (TTA) to dynamically adapt the outlier distribution to the true OOD distribution by utilizing the OOD knowledge embedded in the predicted OOD samples.
This reduces the distribution gap between the outlier and the OOD distributions, enabling a more accurate estimation of OOD scores. As illustrated in Fig. \ref{fig1:subfig2}, a large gap exists between the energy distribution of the outlier data (TinyImages80M \cite{torralba200880}) and the true OOD data (SVHN \cite{netzer2011reading}). By contrast, our adapted outlier distribution is better aligned to the OOD distribution. 
Importantly, the ground truth of the test data is assumed to be unavailable during TTA, and as we will show in the experiments (see Table \ref{ablation_table}), DODA based on the predicted OOD samples can well approximate the upper-bound performance obtained when DODA can get access to the ground truth of the test data to perform TTA (\ie, an oracle model). There have been a few TTA methods for OOD detection, but they require online model retraining~\cite{yang2023auto} or feature memory augmentation~\cite{zhang2023model}. By contrast, DODA focuses on the calibration of the outlier distribution, effectively eliminating the retraining or memory overheads.

On the other hand, training OOD detectors using energy loss functions \cite{liu2020energy,choi2023balanced} is a principled approach to learn the vanilla outlier distribution. However, existing energy losses can underestimate the tail class distribution and involve sensitive hyperparameters on energy margins.  As a result, the vanilla outlier distribution learned by using these losses often misclassifies tail samples as OOD samples during TTA. This can significantly affect the distribution adaptation in DODA, leading to still a relatively large gap between the adapted outlier distribution and the OOD distribution, as shown in Fig. \ref{fig1:subfig2}.
Therefore, AdaptOD introduces a novel Dual-Normalized Energy loss (\textbf{DNE}) to balance energy prediction for imbalanced ID samples and learn a better vanilla outlier distribution for subsequent DODA. Unlike existing energy losses that are focused on sample-wise energy estimation, DNE utilizes both class-wise and sample-wise normalized energy. This helps obtain more balanced prediction energy on the head and tail samples, transferring the energy from the head samples to the tail samples, thereby avoiding the bias toward the head classes (see Fig. \ref{fig1:subfig4}). In doing so, DNE is also free of energy margin hyperparameters and enables the learning of a better vanilla outlier distribution. This guarantees a better starting point for the outlier distribution adaptation and the accuracy of the predicted OOD samples at testing time in DODA, and thus yielding substantially better aligned outlier distribution (see Fig. \ref{fig1:subfig3} vs. Fig. \ref{fig1:subfig2}).
Our main contributions are as follows:
\begin{itemize}
\item We propose the novel approach AdaptOD for OOD detection in LTR. To our best knowledge, it is the first approach for adapting the outlier distribution to the true OOD distribution from both the training and inference stages. 
\item In AdaptOD, we introduce two new components, DODA and DNE, to reduce the gap between the learned outlier distribution and the true OOD distribution in the presence of long-tailed ID data. 
DODA builds upon a vanilla outlier distribution and then dynamically adapts this distribution to the true OOD distribution with the OOD knowledge obtained at testing time. 
DNE is designed to perform class- and sample-wise normalized energy training, which enforces more balanced prediction energy for imbalanced ID samples, enabling the learning of largely enhanced vanilla outlier distribution for more effective DODA.
\item 
Extensive empirical results on three LTR benchmarks CIFAR10-LT, CIFAR100-LT, and ImageNet-LT using six popular OOD datasets demonstrate that AdaptOD substantially outperforms the state-of-the-art (SOTA) OOD detection methods in various LTR scenarios.
\end{itemize}

\section{Related Work}
\paragraph{OOD Detection in Long-Tailed Recognition (LTR).}
In recent years, OOD detection and LTR have been extensively developed. The former determines whether a given input sample belongs to known classes (in-distribution) or unknown classes (out-of-distribution) \cite{sun2021react, wang2023detecting, zhang2023decoupling, wei2022mitigating, tian2022pixel,yu2023block, li2023rethinking,liu2023residual,li2024learning}, while the latter expects to train on class-imbalanced datasets
\cite{tang2022invariant, bai2023effectiveness, shi2023re, alshammari2022long, gou2023rethinking, hong2023long, shi2024long}. 
PASCL \cite{wang2022partial} reveals the difficulty of the OOD detection problem in LTR, and establishes performance benchmarks for OOD detection in LTR based on the SC-OOD benchmark \cite{yang2021semantically}.
This setting is also extended to medical image analysis \cite{mehta2022out, ye2024triaug}, which utilizes a strong data augmentation to discriminate ID data and OOD data.
Recent studies \cite{choi2023balanced, jiang2023detecting} find that fitting the prediction probability of outlier data to a long-tailed distribution is more effective than using a uniform distribution. 
They specify this distribution based on the number of samples in ID classes or a pre-trained ID model to learn this outlier distribution. 
However, it is difficult to obtain such an accurate distribution for outliers in LTR. 
Other studies \cite{wei2023eat, miao2023out} attempt to learn an extra outlier class to overcome the need for learning long-tailed distributions of outliers.
But they need a more complex model design. 
More importantly, all these methods assume that the outlier samples can well represent the distribution of the true OOD data, but this often does not hold in practice since OOD data can be sampled from highly different unknown distributions in different application scenarios. 
Our approach tackles this problem by adapting the outlier distribution to that of the true OOD data.

\paragraph{Test-Time Adaptation (TTA) for OOD Detection.}
Recently, TTA \cite{zhang2022memo, hu2021mixnorm, gao2024atta} has been introduced to OOD detection, in which unlabeled test data that can be seen only once are used to perform online updating pre-trained DNNs for enhancing task performance and quickly adapting to real-world scenarios. 
There are two primary approaches for TTA in other tasks: retraining the model based on unsupervised objectives \cite{wang2020tent, zhang2022domain, wang2022continual} and updating the feature memory for each class \cite{iwasawa2021test, zhang2023adanpc, xiao2022learning}. 
However, unlike these TTA methods that generalize training data to test data and maintain the same label space between them,
TTA for OOD detection \cite{yang2023auto, zhang2023model, fan2022simple} addresses the challenge of identifying unknown classes in test data. 
While training data includes ID data and outlier data, test data comprises not only ID data but also true OOD data consisting of unknown classes that do not overlap with the ID and outlier data.
In particular, AUTO \cite{yang2023auto} is a recent method that attempts to assign pseudo labels to unlabeled test data, and then directly uses these pseudo-labels and test data to retrain the model online through Outlier Exposure \cite{hendrycks2018deep}. 
AdaOOD \cite{zhang2023model} utilizes a memory bank to store feature memories of ID data, then updates these memories online during inference, and lastly employs a $k$NN-based distance method to detect OOD samples. 
However, they fail to work well in the LTR scenarios due to the large variation in the heavily imbalanced training ID data.
Moreover, these methods require retraining or additional memory overheads.
ETLT \cite{fan2022simple} attempts to calibrate OOD scores by a linear regression of its input feature but requires a batch-wise inference to obtain sufficient test samples for the regression.
DODA instead utilizes the dynamically adapted outlier distribution to calibrate the prediction output of test data during inference without any retraining or memory overhead.

\section{Approach}
\paragraph{Preliminaries.}
Let $X^{in}$ denote the input space of the ID data and $Y^{in} = \left\{ 1,2,\dots,k \right\}$ be the set of $k$ imbalanced ID classes in the label space. 
We have genuine OOD data $X^{true\_out}$ that is different from $X^{in}$.
It is normally assumed that genuine OOD data $X^{true\_out}$ are not available during training since OOD samples are unknown instances. 
However, we can obtain auxiliary OOD data from external datasets, which can be used as outliers $X^{aux\_out}$ to act as surrogate OOD data for training/fine-tuning LTR models. That is, $X^{aux\_out}$ is still different from $X^{true\_out}$, but both of them are OOD w.r.t. $X^{in}$.
There is no class overlapping among ID data $X^{in}$, genuine OOD data $X^{true\_out}$, and outlier data $X^{aux\_out}$.
Then the training and test sets can be respectively denoted as: $\mathcal{X}^{train} = X^{in} \cup X^{aux\_out}$ and $\mathcal{X}^{test} = X^{in} \cup X^{true\_out}$. 

OOD detection in LTR is to learn a classifier $f$ with training data $\mathcal{X}^{train}$ so that for any test data $x \in \mathcal{X}^{test}$, if $x$ is drawn from $X^{in}$ (from either head or tail classes), then $f$ can classify $x$ into the correct ID class, whereas if $x$ is drawn from $X^{true\_out}$, then $f$ can detect $x$ as OOD data. 

TTA for OOD detection in LTR is to online update the above pre-trained classifier $f$ with test data $\mathcal{X}^{test}$ during the inference stage, in which for any unlabeled single test sample $x \in \mathcal{X}^{test}$, utilizing pre-trained classifier $f$ to predict whether $x$ belongs to ID or OOD data at the current iteration, then using the predicted label and the test sample $x$ to update the classifier $f$. At the next iteration, the updated classifier $f$ is used to identify a new test sample and continuously update the classifier $f$. 
Notably, each sample can only be seen by $f$ once during inference.

\begin{figure}[t!]
    \centering
    \includegraphics[width=1\columnwidth]{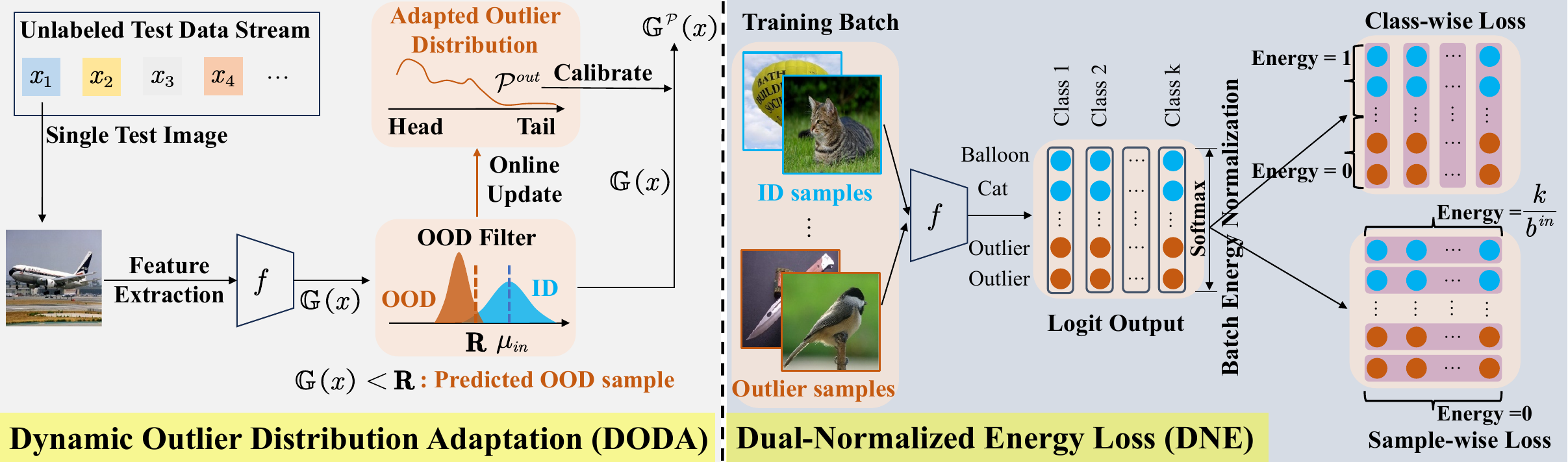}
    \caption{Overview of AdaptOD, which consists of two components, DODA (Left) and DNE (Right). \textbf{Left:} Each test sample is assigned a global energy-based OOD score $\mathbb G(x)$ to adapt the outlier distribution $\mathcal{P}^{out}$. DODA then uses the adapted outlier distribution $\mathcal{P}^{out}$ to calibrate the global energy score $\mathbb G(x)$, obtaining the calibrated global energy score $\mathbb G^{\mathcal{P}}(x)$ as the OOD score.
    \textbf{Right:} For each iteration, DNE first applies Batch Energy Normalization on logit output to obtain the normalized energy, and then utilizes this energy to optimize a dual energy loss function at both the class and sample levels.}
    \label{pipeline}
\end{figure}

\subsection{Overview of AdaptOD}
The proposed AdaptOD approach is designed to tackle the aforementioned distribution shift issue for OOD detection in LTR. As shown in Fig. \ref{pipeline}, AdaptOD consists of two components, namely Dynamic Outlier Distribution Adaptation (DODA) and Dual-Normalized Energy Loss (DNE).
DODA dynamically adapts the learned outlier distribution to the true OOD distribution during inference to reduce the distribution gap between them.
DNE is designed to perform both class-wise and sample-wise normalized energy training to obtain more balanced prediction energy on imbalanced ID samples, thereby yielding an enhanced vanilla outlier distribution and enabling better distribution adaptation in DODA.
Below we introduce each component in detail.

\subsection{DODA: Dynamic Outlier Distribution Adaptation}
Previous OOD detection methods in LTR suffer from a distribution shift between outlier data and true OOD data.
This issue can largely limit the performance of these OOD detectors.
Therefore, we propose to dynamically adapt the outlier distribution to the true OOD distribution and further use it to calibrate the prediction output of test samples at the inference stage.

\paragraph{Dynamic Distribution Adaptation with Predicted OOD Samples.}
Recently, energy-based methods \cite{liu2020energy, choi2023balanced}, which use a global energy score over the ID classes as an OOD score for each test sample, have achieved SOTA performance for OOD detection in LTR.
Motivated by this success, we learn and adapt the  vanilla outlier distribution $\mathcal{P}^{out}$, which is initialized by the global energy from the LTR model predictions on the outlier data, to that of the true OOD data, and then use a $\mathcal{P}^{out}$-calibrated global energy score as the OOD score.
Specifically, given a set of $k$ ID classes, for any test sample $x \in \mathcal{X}^{test}$, its global energy score $\mathbb G(\cdot)$ is defined as \cite{liu2020energy}:
\begin{equation}
\mathbb G(x) = \sum_{j=1}^k e^{f_j(x)}, \label{gx}
\setlength{\abovedisplayskip}{5pt}
\setlength{\belowdisplayskip}{5pt}
\end{equation}
where $f_j(x)$ is the logit output of sample $x$ in class $j, j \in \left\{ 1,2,\dots,k \right\}$.
Let $\mathcal{P}^{out} \in \mathbb R^k$ be an initial outlier distribution.
DODA performs test-time adaptation to dynamically adapt the outlier distribution $\mathcal{P}^{out}$ to the true OOD distribution based on the OOD knowledge from the samples predicted as OOD during inference. 
To this end, we designed an OOD filter using training data to identify OOD samples. Since it is easy to obtain the distribution of global energy score for training ID samples, we use an offline method to determine a threshold for filtering OOD samples based on this energy distribution. This avoids adverse effects on the adaptation speed during inference.
Formally, given ID examples from training data $\mathbf{x} = \{x_1, x_2,..., x_n \}$, where $\mathbf{x} \in X^{in}$ and $n$ is the number of training ID samples, we estimate the mean $\mu_{in}$ and standard deviation $\sigma_{in}$ of the global energy distribution by:
\begin{equation}
\mu_{in} = \frac{\sum_{i=1}^{n}{\mathbb G(x_i)}}{n},
\sigma_{in} = \sqrt{\frac{\sum_{i=1}^{n}{(\mathbb G(x_i)-\mu_{in})^2}}{n-1}}. \label{statistics}
\setlength{\abovedisplayskip}{5pt}
\setlength{\belowdisplayskip}{5pt}
\end{equation}
We then utilize a Z-score-based method to implement the OOD filter, with the Z-score defined as:
\begin{equation}
R = \mu_{in} - \alpha \times \sigma_{in}, \label{filter}
\setlength{\abovedisplayskip}{5pt}
\setlength{\belowdisplayskip}{5pt}
\end{equation}
where $\alpha$ is a hyperparameter. $\alpha=3$ is used by default during the inference stage, and this setting works well throughout our experiments. More discussion about $\alpha$ is described in Appendix \ref{ablation_a}.

To adapt the outlier distribution $\mathcal{P}^{out}$, DODA utilizes the predicted OOD samples by the OOD filter to perform a momentum update of $\mathcal{P}^{out}$ during inference, so that $\mathcal{P}^{out}$ will represent the mean of energy distribution for the predicted OOD samples.
The entries in the vanilla outlier distribution $\mathcal{P}^{out}$ are initialized from the mean energy distribution of the outlier data, and they are updated in an online fashion.
Specifically, when the OOD filter detects the $t$-th test sample $x$ as an OOD sample (\ie, its global energy $\mathbb G(x) < R$), DODA performs an update of $\mathcal{P}^{out}$ as follows: 
\begin{equation}
\mathcal{P}^{out}(t+1) = 
\begin{cases}
\frac{M * \mathcal{P}^{out}(t) + e^{f(x)}}{M+1}, & \mathbb G(x) < R\\
\mathcal{P}^{out}(t), & \mathbb G(x) \geq R
\label{update}
\end{cases}
\setlength{\abovedisplayskip}{5pt}
\setlength{\belowdisplayskip}{5pt}
\end{equation}
where DODA only keep the number of predicted OOD samples $M$ and current $\mathcal{P}^{out}$ during inference. 

\paragraph{Calibrated OOD Score based on the Adapted Outlier Distribution.}
After obtaining the adapted $\mathcal{P}^{out}$, we use it to calibrate the global energy score $\mathbb G(\cdot)$ and define the OOD score as follows:
\begin{equation}
\mathbb G^{\mathcal{P}}(x) = \sum_{j=1}^{k} \frac{e^{f_j(x)}}{1 + \mathcal{P}_j^{out}}, \label{prior}
\setlength{\abovedisplayskip}{5pt}
\setlength{\belowdisplayskip}{5pt}
\end{equation}
where $x \in \mathcal{X}^{test}$ and $\mathbb G^{\mathcal{P}}(\cdot)$ denotes the calibrated global energy score with the adapted outlier distribution. This way helps reduce the energy proportion of the head classes that true OOD distribution leans toward in the original global energy score $\mathbb G(\cdot)$.
In doing so, the distribution gap between the outliers and true OOD is effectively reduced in final OOD score $\mathbb G^{\mathcal{P}}(\cdot)$, enabling more accurate estimation of OOD scores in heavily imbalanced ID data without incurring any retraining cost or additional memory expense.

\subsection{DNE: Dual-Normalized Energy Loss}
When using existing global energy loss to obtain the vanilla outlier distribution $\mathcal{P}^{out}$, the distribution of tail samples is indistinguishable from that of OOD samples due to the underestimating of tail samples. 
We are also required to manually tune the sensitive hyperparameters on energy margins under complex class imbalance. These can lead an inaccurate OOD filter used in Eq. \ref{filter} and subsequently affect the distribution adaptation in DODA.
To tackle these issues, we propose a Dual-Normalized Energy Loss (DNE), which consists of two novel components, namely class-wise normalized energy loss (DNE-C) and sample-wise normalized energy loss (DNE-S).
DNE-C is a class-wise training loss for balancing the sum of energy on all ID samples for each ID class, whereas DNE-S is a sample-wise training loss for balancing the sum of energy on all ID classes for each ID sample. 
DNE learns a balanced prediction energy distribution on imbalanced ID samples, which helps further reduce the bias toward the head classes in Eq. \ref{prior},
thereby improving vanilla outlier distribution for a better OOD filter in Eq. \ref{filter} and a better vanilla outlier distribution $\mathcal{P}^{out}$ in Eq. \ref{update}. 
It also provides stable energy margins, eliminating the need of manual tuning of these margins.

\paragraph{Batch Energy Normalization.}
To this end, we first propose a novel batch energy normalization method, which conducts energy normalization on the logit output of each class for a batch of training samples. In doing so, the energy of each sample is dependent on the energy of other ID samples and OOD samples relative to the same class. This helps transfer the energy knowledge from the head samples to the tail samples, enabling a better estimation for the energy distribution of tail samples.

Formally, let $\mathbf{x}^{in} \in X^{in}$ be one training batch of ID data, with $\mathbf{x}^{in} = \{ x_1^{in}, x_2^{in},..., x_{b^{in}}^{in} \}$ and $b^{in}$ be its batch size, and $\mathbf{x}^{out} \in X^{aux\_out}$ be a set of outlier data in a training batch, with $ \mathbf{x}^{out} = \{ x_1^{out}, x_2^{out},..., x_{b^{out}}^{out} \}$ whose set size is $b^{out}$, then the batch energy normalization $F_j(x_i)$ for a sample $x_i \in x^{in} \cup x^{out}$ in class $j \in \{1,2,...,k\}$  with classifier $f$ is defined as: 
\begin{equation}
F_j(x_i) = \frac{e^{f_j(x_i)}}{e^{f_j(x_1^{in})}+...+e^{f_j(x_{b^{in}}^{in})}+e^{f_j(x_{1}^{out})}+...+e^{f_j(x_{b^{out}}^{out})}}, \label{Normalize}
\setlength{\abovedisplayskip}{5pt}
\setlength{\belowdisplayskip}{5pt}
\end{equation}
where $f_j(x)$ is the logit output of sample $x$ in class $j$. 
Essentially, we use the logit output of all samples in a training batch on a class $j$ to normalize the energy prediction of sample $x$. 
This largely reduces the energy prediction bias toward the head samples. The energy of the outlier data is included as a calibration modulation. 
Then, those normalized energy scores are used for the dual-normalized energy losses, DNE-C and DNE-S, to better balance the prediction energy of long-tailed ID samples.

Additionally, compared to the current energy-based method \cite{choi2023balanced, liu2020energy} for OOD detection in LTR that requires manually designed energy margin hyperparameters,
batch energy normalization adjusts the energy of the batch samples on each class to the same scale, so it can provide stable energy margins for the balanced training without relying on the training dataset and/or the class imbalance factor, without the need of manually tuning them. 

\paragraph{Class-wise Normalized Energy Loss (DNE-C). }
DNE-C independently regularizes the energy for each class to enhance the normalized energy of ID samples for more class-wise balanced energy. Formally, let $\mathcal {D}_{in} = (X^{in}, Y^{in})$ and $\mathcal {D}_{out} = X^{aux\_out}$, then we can independently minimize the class energy on each class as follows:
\begin{equation}
\mathcal {L}_{C} = \sum_{j=1}^k (\mathbb {E}_{(\mathbf{x},y) \sim \mathcal {D}_{in}}[(max(0,m_{in}^c - \mathbf {C}_j (\mathbf{x})))^2]
+\mathbb {E}_{\mathbf{x} \sim \mathcal {D}_{out}}[(max(0,\mathbf {C}_j (\mathbf{x})-m_{out}^c))^2]), \label{CW-loss}
\setlength{\abovedisplayskip}{5pt}
\setlength{\belowdisplayskip}{5pt}
\end{equation}
where $m_{in}^c=1$ and $m_{out}^c=0$ are the default margin hyperparameter settings without the need of manual tuning on different datasets (see Appendix \ref{margin} for more details). The class-wise normalized energy $\mathbf {C}_j (\mathbf{x}), j \in \{1,2,...,k\}$ is defined as:
\begin{equation}
\mathbf {C}_j (\mathbf{x}) = \sum_{i=1}^{b} F_j(x_i), \label{C-loss}
\setlength{\abovedisplayskip}{5pt}
\setlength{\belowdisplayskip}{5pt}
\end{equation}
where $b$ is the batch size of the batch $\mathbf{x}$ (if $\mathbf{x}$ is  $\mathbf{x}^{in} $ that $b$ is $ b^{in}$,  and $\mathbf{x}$ is $\mathbf{x}^{out} $ that $b$ is $ b^{out}$), and $x_i$ is the $i$-th sample in the batch $\mathbf{x}$.
Notably, even if some classes do not have the corresponding ID samples in a certain training batch, this loss also can work well. 
This is because there is less distribution shift among classes in the ID data compared to the OOD data.
Therefore, the output of ID samples on incorrect ID classes should also be higher than the OOD samples.
DNE-C balances the sum of energy on all ID samples for each ID class and distinguishes outlier samples from ID samples, especially for the underestimated tail classes in a class-wise manner.

\paragraph{Sample-wise Normalized Energy Loss (DNE-S). }
DNE-S independently regularizes the energy for each sample to enhance the energy of ID samples for sample-wise balanced energy.
Formally, we minimize the global energy over all classes of each sample as follows:
\begin{equation}
\mathcal {L}_{S} = \mathbb {E}_{(x,y) \sim \mathcal {D}_{in}}[(max(0,m_{in}^s - \mathbf {S} (x)))^2]
+\mathbb {E}_{{x} \sim \mathcal {D}_{out}}[(max(0,\mathbf {S} (x)-m_{out}^s))^2], \label{SW-loss}
\setlength{\abovedisplayskip}{5pt}
\setlength{\belowdisplayskip}{5pt}
\end{equation}
where $x \in \mathbf{x}^{in} \cup \mathbf{x}^{out}$, $m_{in}^s=\frac{k}{b^{in}}$ and $m_{out}^s=0 $ are the default margin hyperparameter settings that can also work stably regardless of the ID/OOD datasets (see Appendix \ref{margin}). Then the sample-wise normalized energy $\mathbf {S} (x)$ can be defined as:
\begin{equation}
\mathbf {S} (x) = \sum_{j=1}^{k} F_j(x). \label{S-loss}
\setlength{\abovedisplayskip}{5pt}
\setlength{\belowdisplayskip}{5pt}
\end{equation}
After doing this, we can regularize the global energy of the ID data, particularly the low global energy for tail samples. 
DNE-S efficiently balances the energy between head and tail samples.
As a result, the combination of DNE-C and DNE-S can learn substantially more balanced prediction energy of ID samples, facilitating DODA to solve the distribution shift problems.

\noindent\textbf{Overall Training Objective.} Overall, we utilize the cross-entropy loss, together with our two normalized energy losses, to train our model. The final objective of our DNE training is as follows:
\begin{equation}
\mathcal {L}_{total} = \mathbb {E}_{x,y \sim \mathcal {D}_{in}}[\ell(f(x),y] + \mathcal {L}_{dne}, \label{total}
\setlength{\abovedisplayskip}{5pt}
\setlength{\belowdisplayskip}{5pt}
\end{equation}
where $\ell$ is a cross-entropy loss, along with the two normalized energy losses:
\begin{small}
\begin{align}
\begin{aligned}
\mathcal {L}_{dne} = 
& \mathcal {L}_{S} + \mathcal {L}_{C} =
 \mathbb {E}_{(x,y) \sim \mathcal {D}_{in}}[(max(0,m_{in}^s - \mathbf {S} (x)))^2]
+\mathbb {E}_{{x} \sim \mathcal {D}_{out}}[(max(0,\mathbf {S} (x)-m_{out}^s))^2], \label{total-fine} \\
& + \sum_{j=1}^k (\mathbb {E}_{(\mathbf{x},y) \sim \mathcal {D}_{in}}[(max(0,m_{in}^c - \mathbf {C}_j (\mathbf{x})))^2] 
+\mathbb {E}_{\mathbf{x} \sim \mathcal {D}_{out}}[(max(0,\mathbf {C}_j (\mathbf{x})-m_{out}^c))^2]) 
\setlength{\abovedisplayskip}{5pt}
\setlength{\belowdisplayskip}{5pt}
\end{aligned}
\end{align}
\end{small}
where $\mathcal {L}_{C}$ is as defined in Eq.~\ref{CW-loss} and $\mathcal {L}_{S}$ is as defined in Eq.~\ref{SW-loss}. The algorithm of AdaptOD described in Appendix \ref{Algo}.

\begin{table}[t!]
\centering
\large
\caption{Comparison of AdaptOD with EnergyOE and COCL on six OOD datasets.}
\label{fine-table}
\scalebox{0.7}{
\begin{tabular}{c|c|cccc|cccc}
\hline
\multirow{2}{*}{\shortstack{OOD \\ Dataset}} & \multirow{2}{*}{Method} & \multicolumn{4}{c|}{ID Dataset: CIFAR10-LT} & \multicolumn{4}{c}{ID Dataset: CIFAR100-LT} \\ \cline{3-10}
 & & AUC$\uparrow$ & AP-in$\uparrow$ & AP-out$\uparrow$ & FPR$\downarrow$ & AUC$\uparrow$ & AP-in$\uparrow$ & AP-out$\uparrow$ & FPR$\downarrow$ \\ 
\hline
\multirow{3}{*}{Texture\cite{cimpoi2014describing}}
 & EnergyOE\cite{liu2020energy} & 95.53 & 97.42 & 92.93 & 18.44 & 79.56 & 86.03 & 70.88 & 79.45 \\ 
 & COCL\cite{miao2023out} & 96.81 & 98.21 & 93.86 & 14.65 & 81.99 & 88.05 & 74.38 & 59.79 \\ 
 & \textbf{AdaptOD(Ours)} & \textbf{98.22} & \textbf{98.81} & \textbf{94.91} & \textbf{11.60} & \textbf{83.88} & \textbf{89.43} & \textbf{76.47} & \textbf{58.47} \\
\hline
\multirow{3}{*}{SVHN\cite{netzer2011reading}}
 & EnergyOE\cite{liu2020energy} & 96.63 & 92.33 & 98.46 & 14.37 & 86.19 & 81.42 & 91.74 & 34.36 \\ 
 & COCL\cite{miao2023out} & 96.98 & 93.25 & 98.61 & 12.59 & 89.20 & 81.57 & 94.21 & 54.46 \\ 
 & \textbf{AdaptOD(Ours)} & \textbf{98.13} & \textbf{94.34} & \textbf{99.11} & \textbf{10.33} & \textbf{93.09} & \textbf{91.32} & \textbf{96.86} & \textbf{17.63} \\
\hline
\multirow{3}{*}{CIFAR\cite{krizhevsky2009learning}}
 & EnergyOE\cite{liu2020energy} & 84.44 & 85.74 & 84.63 & 61.73 &  61.15 & 67.12 & 56.66 & 91.42 \\ 
 & COCL\cite{miao2023out} & 86.63 & 86.66 & 86.28 & 52.21 & 62.05 & 66.14 & 56.82 & 93.88 \\
 & \textbf{AdaptOD(Ours)} & \textbf{89.05} & \textbf{89.93} & \textbf{88.22} & \textbf{45.51} & \textbf{72.77} & \textbf{76.37} & \textbf{70.58} & \textbf{86.04} \\
\hline
\multirow{3}{*}{\shortstack{TIN~\cite{le2015tiny}}}
 & EnergyOE\cite{liu2020energy} & 88.40 & 91.65 & 84.95 & 46.23 &  70.78 & 79.40 & 55.90 & 90.74 \\ 
 & COCL\cite{miao2023out} & 90.43 & 92.52 & 87.03 & 46.12 & 71.87 & 81.89 & 57.12 & 83.93 \\
 & \textbf{AdaptOD(Ours)} & \textbf{91.40} & \textbf{93.85} & \textbf{88.18} & \textbf{42.77} & \textbf{72.87} & \textbf{82.06} & \textbf{58.92} & \textbf{88.24} \\
\hline
\multirow{3}{*}{LSUN\cite{yu2015lsun}}
 & EnergyOE\cite{liu2020energy} & 94.00 & 94.78 & 93.70 & 28.42 &  81.61 & 86.57 & 69.16 &  80.57\\ 
 & COCL\cite{miao2023out} & 94.85 & 95.43 & 93.98 & 27.48 & 84.10 & 89.89 & 69.80 & 74.67 \\
 & \textbf{AdaptOD(Ours)} & \textbf{96.16} & \textbf{96.84} & \textbf{95.86} & \textbf{24.12} & \textbf{85.70} & \textbf{90.55} & \textbf{72.70} & \textbf{70.20} \\
\hline
\multirow{3}{*}{Place365\cite{zhou2017places}}
 & EnergyOE\cite{liu2020energy} & 92.51 & 84.26 & 97.14 & 33.63 &  79.12 & 63.38 & 89.09 &  81.43\\ 
 & COCL\cite{miao2023out} & 93.97 & 87.36 & 97.56 & 32.25 & 80.30 & 68.65 & 89.16 & 77.83 \\
 & \textbf{AdaptOD(Ours)} & \textbf{95.19} & \textbf{89.56} & \textbf{98.44} & \textbf{29.22} & \textbf{83.27} & \textbf{68.82} & \textbf{91.44} & \textbf{71.63} \\
\hline
\end{tabular}}
\end{table}

\begin{table}[t!]
\centering
\large
\caption{\centering{Comparison to different long-tailed OOD detection methods.}}
\label{CIFAR-OOD}
\scalebox{0.65}{
\begin{tabular}{c|ccccc|ccccc}
\hline
\multirow{2}{*}{Method}  & \multicolumn{5}{c|}{ID Dataset: CIFAR10-LT} & \multicolumn{5}{c}{ID Dataset: CIFAR100-LT} \\ \cline{2-11}
 & AUC$\uparrow$ & AP-in$\uparrow$ & AP-out$\uparrow$ & FPR$\downarrow$ & ACC$\uparrow$ & AUC$\uparrow$ & AP-in$\uparrow$ & AP-out$\uparrow$ & FPR$\downarrow$  & ACC$\uparrow$ \\ 
\hline
OE\cite{hendrycks2018deep} & 89.76 & 89.45 & 87.22 & 53.19 & 73.59   &   73.52 & 75.06 & 67.27 & 86.30 & 39.42\\
EnergyOE\cite{liu2020energy} & 91.92 & 91.03 & 91.97 & 33.80 & 74.57   &   76.40 & 77.32 & 72.24 & 76.33 & 41.32\\
PASCL\cite{wang2022partial} & 90.99 & 90.56 & 89.24 & 42.90 & 77.08   &   73.32 & 74.84 & 67.18 & 79.38 & 43.10\\
EAT\cite{wei2023eat} & 92.87 & 91.76 & 92.40 & 32.42 & 81.31   &   75.45 & 76.02 & 70.87 & 77.83 & 46.23\\
Class Prior\cite{jiang2023detecting} & 92.08 & 91.17 & 90.86 & 34.42 & 74.33   &   76.03 & 77.31 & 72.26 & 76.43 & 40.77\\
BERL\cite{choi2023balanced} & 92.56 & 91.41 & 91.94 & 32.83 & 81.37   &   77.75 & 78.61 & 73.10 & 74.86 & 45.88\\
COCL\cite{miao2023out} & 93.28 & 92.24 & 92.89 & 30.88 & 81.56    &   78.25 & 79.37 & 73.58 & 74.09 & 46.41\\
\hline
OE\cite{hendrycks2018deep}+\textbf{DODA(Ours)} & 91.62 & 90.55 & 89.39 & 49.02 & 73.59   &   75.46   & 77.14 & 69.88 & 83.67 & 39.42\\
EnergyOE\cite{liu2020energy}+\textbf{DODA(Ours)} & 93.36 & 92.17 & 92.97 & 30.82 & 74.57   &   79.40 & 80.89 & 76.54 & 72.63 & 41.32\\
BERL\cite{choi2023balanced}+\textbf{DODA(Ours)} & 93.77 & 92.62 & 93.15 & 29.41 & 81.37   &   79.45 & 81.15 & 75.52 & 70.51 & 45.88\\
COCL\cite{miao2023out}+\textbf{DODA(Ours)} & 93.89 & 93.06 & 93.39 & 29.25 & 81.56    &   79.81 & 81.26 & 75.93 & 70.14 & 46.41\\
\hline
\textbf{AdaptOD(Ours)} & \textbf{94.69} & \textbf{93.89} & \textbf{94.12} & \textbf{27.26} & \textbf{82.27}    &   \textbf{81.93} & \textbf{83.09} & \textbf{77.83} & \textbf{67.37} & \textbf{47.91}\\
\hline
\end{tabular}}
\end{table}

\begin{table}[t!]
\centering
\large
\caption{\centering{Comparison to different TTA-based OOD detection methods.}}
\label{CIFAR-TTA}
\scalebox{0.63}{
\begin{tabular}{cc|cccc|cccc}
\hline
\multirow{2}{*}{\shortstack{Training \\ Method}} & \multirow{2}{*}{\shortstack{TTA \\ Method}} & \multicolumn{4}{c|}{ID Dataset: CIFAR10-LT} & \multicolumn{4}{c}{ID Dataset: CIFAR100-LT} \\ \cline{3-10}
 & & AUC$\uparrow$ & AP-in$\uparrow$ & AP-out$\uparrow$ & FPR$\downarrow$ & AUC$\uparrow$ & AP-in$\uparrow$ & AP-out$\uparrow$ & FPR$\downarrow$ \\ 
\hline
\multirow{4}{*}{OE\cite{hendrycks2018deep}} & w/o TTA & 89.76{\scriptsize $\pm$0.27} & 89.45{\scriptsize $\pm$0.56} & 87.22{\scriptsize $\pm$0.61} & 53.19{\scriptsize $\pm$0.42} &   73.52{\scriptsize $\pm$0.68} & 75.06{\scriptsize $\pm$0.59} & 67.27{\scriptsize $\pm$0.57} & 86.30{\scriptsize $\pm$0.92}  \\
 & AUTO\cite{yang2023auto} &  90.49{\scriptsize $\pm$0.29} & 89.83{\scriptsize $\pm$0.52} & 87.45{\scriptsize $\pm$0.83} & 52.63{\scriptsize $\pm$0.47} &  73.93{\scriptsize $\pm$0.89} & 75.98{\scriptsize $\pm$0.81} & 67.74{\scriptsize $\pm$0.65} & 85.71{\scriptsize $\pm$1.00} \\
 & AdaODD\cite{zhang2023model} & 90.89{\scriptsize $\pm$0.26} & 90.17{\scriptsize $\pm$0.51} & 87.88{\scriptsize $\pm$0.84} & 51.44{\scriptsize $\pm$0.56} & 74.67{\scriptsize $\pm$0.92} & 76.53{\scriptsize $\pm$0.64} & 67.89{\scriptsize $\pm$0.82} & 85.34{\scriptsize $\pm$0.94}  \\
 & \textbf{DODA(Ours)} & \underline{91.62{\scriptsize $\pm$0.23}} & \underline{90.55{\scriptsize $\pm$0.45}} & \underline{89.39{\scriptsize $\pm$0.68}} & \underline{49.02{\scriptsize $\pm$0.41}} &  \underline{75.46{\scriptsize $\pm$0.77}} & \underline{77.14{\scriptsize $\pm$0.59}} & \underline{69.88{\scriptsize $\pm$0.80}} & \underline{83.67{\scriptsize $\pm$0.88}}  \\
\hline
\multirow{4}{*}{EnergyOE\cite{liu2020energy}} & w/o TTA & 91.92{\scriptsize $\pm$0.30} & 91.03{\scriptsize $\pm$0.53} & 91.97{\scriptsize $\pm$0.62} & 33.80{\scriptsize $\pm$0.56} &   76.40{\scriptsize $\pm$0.86} & 77.32{\scriptsize $\pm$0.59} & 72.24{\scriptsize $\pm$0.62} & 76.33{\scriptsize $\pm$1.03}  \\
 & AUTO\cite{yang2023auto} & 92.48{\scriptsize $\pm$0.32} & 91.43{\scriptsize $\pm$0.55} & 92.44{\scriptsize $\pm$0.79} & 31.99{\scriptsize $\pm$0.36} &   77.65{\scriptsize $\pm$1.01} & 78.11{\scriptsize $\pm$0.62} & 74.18{\scriptsize $\pm$0.78} & 74.66{\scriptsize $\pm$0.99}  \\
 & AdaODD\cite{zhang2023model} & 92.28{\scriptsize $\pm$0.26} & 91.63{\scriptsize $\pm$0.56} & 91.73{\scriptsize $\pm$0.61} & 32.83{\scriptsize $\pm$0.59} &  77.67{\scriptsize $\pm$0.82} & 78.47{\scriptsize $\pm$0.81} & 74.05{\scriptsize $\pm$0.83} & 74.86{\scriptsize $\pm$0.98}  \\
 & \textbf{DODA(Ours)} & \underline{93.36{\scriptsize $\pm$0.28}} & \underline{92.17{\scriptsize $\pm$0.53}} & \underline{92.97{\scriptsize $\pm$0.70}} & \underline{30.82{\scriptsize $\pm$0.51}} & \underline{79.40{\scriptsize $\pm$0.98}} & \underline{80.89{\scriptsize $\pm$0.84}} & \underline{76.54{\scriptsize $\pm$0.64}} & \underline{72.63{\scriptsize $\pm$0.94}}   \\
\hline
\multirow{4}{*}{BERL\cite{choi2023balanced}} & w/o TTA & 92.56{\scriptsize $\pm$0.40} & 91.41{\scriptsize $\pm$0.83} & 91.94{\scriptsize $\pm$0.85} & 32.83{\scriptsize $\pm$0.38} &   77.75{\scriptsize $\pm$0.77} & 78.61{\scriptsize $\pm$0.56} & 73.10{\scriptsize $\pm$0.73} & 74.86{\scriptsize $\pm$1.07} \\
 & AUTO\cite{yang2023auto} & 92.41{\scriptsize $\pm$0.49} & 91.73{\scriptsize $\pm$0.56} & 92.42{\scriptsize $\pm$0.90} & 31.91{\scriptsize $\pm$0.36} &  77.99{\scriptsize $\pm$0.75} & 78.50{\scriptsize $\pm$0.84} & 73.50{\scriptsize $\pm$0.87} & 74.03{\scriptsize $\pm$1.00}  \\
 & AdaODD\cite{zhang2023model} & 92.68{\scriptsize $\pm$0.26} & 91.79{\scriptsize $\pm$0.54} & 92.20{\scriptsize $\pm$0.67} & 31.41{\scriptsize $\pm$0.51} &  78.26{\scriptsize $\pm$0.97} & 78.94{\scriptsize $\pm$0.81} & 73.61{\scriptsize $\pm$0.75} & 73.76{\scriptsize $\pm$1.12}  \\
 & \textbf{DODA(Ours)} & \underline{93.77{\scriptsize $\pm$0.30}} & \underline{92.62{\scriptsize $\pm$0.51}} & \underline{93.15{\scriptsize $\pm$0.73}} & \underline{29.41{\scriptsize $\pm$0.37}} &  \underline{79.45{\scriptsize $\pm$0.83}} & \underline{81.15{\scriptsize $\pm$0.79}} & \underline{75.52{\scriptsize $\pm$0.69}} & \underline{70.51{\scriptsize $\pm$0.91}}  \\
\hline
\multirow{4}{*}{COCL\cite{miao2023out}} & w/o TTA & 93.28{\scriptsize $\pm$0.30} & 92.24{\scriptsize $\pm$0.78} & 92.89{\scriptsize $\pm$0.72} & 30.88{\scriptsize $\pm$0.63} &   78.25{\scriptsize $\pm$0.99} & 79.37{\scriptsize $\pm$0.65} & 73.58{\scriptsize $\pm$0.76} & 74.09{\scriptsize $\pm$0.85}\\
 & AUTO\cite{yang2023auto} & 93.62{\scriptsize $\pm$0.43} & 92.74{\scriptsize $\pm$0.83} & 93.10{\scriptsize $\pm$0.59} & 30.41{\scriptsize $\pm$0.40} &   78.85{\scriptsize $\pm$0.97} & 79.99{\scriptsize $\pm$0.72} & 74.01{\scriptsize $\pm$0.86} & 72.75{\scriptsize $\pm$0.95}  \\
 & AdaODD\cite{zhang2023model} & 93.48{\scriptsize $\pm$0.22} & 92.60{\scriptsize $\pm$0.66} & 93.05{\scriptsize $\pm$0.81} & 30.79{\scriptsize $\pm$0.39} &  79.07{\scriptsize $\pm$0.70} & 80.00{\scriptsize $\pm$0.60} & 74.60{\scriptsize $\pm$0.84} & 73.09{\scriptsize $\pm$0.91} \\
 & \textbf{DODA(Ours)} & \underline{93.89{\scriptsize $\pm$0.36}} & \underline{93.06{\scriptsize $\pm$0.56}} & \underline{93.39{\scriptsize $\pm$0.74}} & \underline{29.25{\scriptsize $\pm$0.40}} &  \underline{79.81{\scriptsize $\pm$0.96}} & \underline{81.26{\scriptsize $\pm$0.59}} & \underline{75.93{\scriptsize $\pm$0.72}} & \underline{70.14{\scriptsize $\pm$0.98}}\\
\hline
\multirow{4}{*}{\textbf{\shortstack{DNE \\ (Ours)}}} & w/o TTA & 92.77{\scriptsize $\pm$0.48} & 92.18{\scriptsize $\pm$0.71} & 92.62{\scriptsize $\pm$0.61} & 31.48{\scriptsize $\pm$0.36} & 77.92{\scriptsize $\pm$0.75} & 78.97{\scriptsize $\pm$0.61} & 73.92{\scriptsize $\pm$0.81} & 74.44{\scriptsize $\pm$0.99} \\
 & AUTO\cite{yang2023auto} & 92.89{\scriptsize $\pm$0.44} & 92.69{\scriptsize $\pm$0.86} & 92.25{\scriptsize $\pm$0.60} & 30.85{\scriptsize $\pm$0.62} &  79.36{\scriptsize $\pm$0.91} & 80.19{\scriptsize $\pm$0.63} & 74.81{\scriptsize $\pm$0.80} & 72.10{\scriptsize $\pm$1.19}  \\
 & AdaODD\cite{zhang2023model} & 93.39{\scriptsize $\pm$0.46} & 92.27{\scriptsize $\pm$0.69} & 92.92{\scriptsize $\pm$0.59} & 30.78{\scriptsize $\pm$0.55} &  80.26{\scriptsize $\pm$0.81} & 81.72{\scriptsize $\pm$0.68} & 75.62{\scriptsize $\pm$0.88} & 71.96{\scriptsize $\pm$0.95} \\
 & \textbf{DODA(Ours)} & \textbf{94.69{\scriptsize $\pm$0.22}} & \textbf{93.89{\scriptsize $\pm$0.68}} & \textbf{94.12{\scriptsize $\pm$0.58}} & \textbf{27.26{\scriptsize $\pm$0.49}} & \textbf{81.93{\scriptsize $\pm$0.71}} & \textbf{83.09{\scriptsize $\pm$0.64}} & \textbf{77.83{\scriptsize $\pm$0.76}} & \textbf{67.37{\scriptsize $\pm$0.93}}\\
\hline
\end{tabular}}
\end{table}

\section{Experiments}

\subsection{Experiment Settings}
\paragraph{Datasets.}
Following \cite{wang2022partial,miao2023out, wei2023eat}, we use three popular long-tailed datasets CIFAR10-LT \cite{cao2019learning}, CIFAR100-LT \cite{cao2019learning} and ImageNet-LT \cite{liu2019large} as ID data $X^{in}$. 
The default imbalance ratio is set to $\rho=100$ on CIFAR10/100-LT.
TinyImages80M \cite{torralba200880} is used as the outlier data $X^{aux\_out}$ for CIFAR10/100-LT and ImageNet-Extra \cite{wang2022partial} is used as outlier data for ImageNet-LT. We use six datasets CIFAR \cite{krizhevsky2009learning}, Texture \cite{cimpoi2014describing}, SVHN \cite{netzer2011reading}, LSUN \cite{yu2015lsun}, Places365 \cite{zhou2017places} and TinyImageNet \cite{le2015tiny}, all of which are introduced in the SC-OOD benchmark \cite{yang2021semantically} as the OOD test set for CIFAR10/100-LT, and ImageNet-1k-OOD \cite{wang2022partial} as the OOD test set for ImageNet-LT. More details about the datasets are presented in Appendix \ref{datasets}.

\paragraph{Implementation Details.}
Our AdaptOD is compared with seven SOTA OOD detection methods on long-tailed data, including two popular methods: OE \cite{hendrycks2018deep} and EnergyOE \cite{liu2020energy}, and five recent methods: PASCL \cite{wang2022partial}, EAT \cite{wei2023eat}, Class Prior \cite{jiang2023detecting}, BERL \cite{choi2023balanced}, and COCL \cite{miao2023out}. 
Further, we also compare AdaptOD with two SOTA TTA methods for OOD detection, including AUTO \cite{yang2023auto} and AdaOOD \cite{zhang2023model}. 
We use ResNet18 \cite{he2016deep} as our backbone on CIFAR10/100-LT and ResNet50 \cite{he2016deep} on ImageNet-LT.
Following fine-tuning-based methods OE \cite{hendrycks2018deep}, EnergyOE \cite{liu2020energy}, and BERL \cite{choi2023balanced}, our approach AdaptOD employs a similar training strategy to them that we obtain a pre-trained model with only ID data and fine-tune this model with both ID data and outlier data.
The reported results are averaged over six independent runs. More details about the implementation details are presented in Appendix \ref{Implement}.

\paragraph{Evaluation Measures.}
Following \cite{yang2021semantically, miao2023out}, we use the below common metrics for OOD detection and ID classification: 
(1) FPR is the false positive rate of OOD examples when the true positive rate of ID examples is at 95\%,
(2) AUC computes the area under the receiver operating characteristic curve of detecting OOD samples,
(3) AP measures the area under the precision-recall curve, which can be either AP-in in which ID samples are treated as positive or AP-out in which OOD samples are regarded as positive,
and (4) ACC calculates the classification accuracy of the long-tailed ID data. The reported results are averaged over six independent runs with different random seeds by default.

\begin{table}[t!]
\caption{Comparison results on the large-scale ID dataset ImageNet-LT.} 
\label{ImageNet}
\large
\begin{subtable}{0.46\linewidth}
\scalebox{0.55}{
\begin{tabular}{c|c|c|c|c|c}
\hline
Method & AUC$\uparrow$ & AP-in$\uparrow$ & AP-out$\uparrow$ & FPR$\downarrow$ & ACC$\uparrow$ \\ 
\hline
OE\cite{hendrycks2018deep} & 68.33 & 43.87 & 82.54 & 90.98 & 44.00\\
EnergyOE\cite{liu2020energy} & 69.43 & 45.12 & 84.75 & 76.89 & 44.42 \\
EAT\cite{wei2023eat} & 69.84 & 43.15 & 81.32 & 80.97 & 46.79 \\
PASCL\cite{wang2022partial} & 68.00 & 43.32 & 82.69 & 82.28 & 47.29\\
Class Prior\cite{jiang2023detecting} & 70.43 & 45.26 & 84.82 & 77.63 & 46.83 \\
BERL\cite{choi2023balanced} & 71.16 & 45.97 & 85.63 & 76.98 & 50.42 \\
COCL\cite{miao2023out} & 71.85 & 46.76 & 86.21 & 75.60 & 51.11 \\
\hline
BERL\cite{choi2023balanced}+\textbf{DODA} & 73.12 & 47.34 & 86.95 & 74.92 & 50.42 \\
COCL\cite{miao2023out}+\textbf{DODA} & 73.27 & 47.98 & 87.77 & 74.71 & 51.11 \\
\hline
\textbf{AdaptOD(Ours)} & \textbf{74.32} & \textbf{49.02} & \textbf{88.63} & \textbf{72.91}  & \textbf{51.67} \\
\hline
\end{tabular}}
\end{subtable}
\begin{subtable}{0.44\linewidth}
\scalebox{0.54}{
\begin{tabular}{cc|cccc}
\hline
Training & Test & AUC$\uparrow$ & AP-in$\uparrow$ & AP-out$\uparrow$ & FPR$\downarrow$\\ 
\hline
\multirow{4}{*}{BERL\cite{choi2023balanced}} & w/o TTA & 71.16{\scriptsize $\pm$0.96} & 45.97{\scriptsize $\pm$0.85} & 85.63{\scriptsize $\pm$0.77} & 76.98{\scriptsize $\pm$1.79} \\
 & AUTO\cite{yang2023auto} & 71.66{\scriptsize $\pm$1.20} & 46.58{\scriptsize $\pm$0.80} & 86.05{\scriptsize $\pm$0.77} & 76.09{\scriptsize $\pm$1.63}  \\
 & AdaODD\cite{zhang2023model} & 71.80{\scriptsize $\pm$1.14} & 46.47{\scriptsize $\pm$0.63} & 85.56{\scriptsize $\pm$1.01} & 77.36{\scriptsize $\pm$1.69}  \\
 & \textbf{DODA(Ours)} & \underline{73.12{\scriptsize $\pm$1.18}} & \underline{47.34{\scriptsize $\pm$0.75}} & \underline{86.95{\scriptsize $\pm$0.76}} & \underline{74.92{\scriptsize $\pm$1.67}}   \\
\hline
\multirow{4}{*}{COCL\cite{miao2023out}} &  w/o TTA  & 71.85{\scriptsize $\pm$1.15} & 46.76{\scriptsize $\pm$1.13} & 86.21{\scriptsize $\pm$1.11} & 75.60{\scriptsize $\pm$1.38}  \\
 & AUTO\cite{yang2023auto} & 71.79{\scriptsize $\pm$1.22} & 46.84{\scriptsize $\pm$0.81} & 86.89{\scriptsize $\pm$1.18} & 75.28{\scriptsize $\pm$1.69}    \\
 & AdaODD\cite{zhang2023model} & 72.35{\scriptsize $\pm$1.10} & 47.20{\scriptsize $\pm$1.16} & 86.89{\scriptsize $\pm$0.96} & 75.06{\scriptsize $\pm$1.91}  \\
 & \textbf{DODA(Ours)} & \underline{73.27{\scriptsize $\pm$1.19}} & \underline{47.98{\scriptsize $\pm$1.00}} & \underline{87.77{\scriptsize $\pm$0.74}} & \underline{74.71{\scriptsize $\pm$1.55}}  \\
\hline
\multirow{4}{*}{\textbf{\shortstack{DNE \\ (Ours)}}} &  w/o TTA  & 72.04{\scriptsize $\pm$1.07} & 46.53{\scriptsize $\pm$0.72} & 86.06{\scriptsize $\pm$0.78} & 75.82{\scriptsize $\pm$1.38}   \\
 & AUTO\cite{yang2023auto} & 73.31{\scriptsize $\pm$1.26} & 47.26{\scriptsize $\pm$1.14} & 87.11{\scriptsize $\pm$1.19} & 74.60{\scriptsize $\pm$1.27}  \\
 & AdaODD\cite{zhang2023model} & 73.10{\scriptsize $\pm$0.81} & 46.83{\scriptsize $\pm$0.76} & 86.68{\scriptsize $\pm$0.74} & 74.64{\scriptsize $\pm$1.41}   \\
 & \textbf{DODA(Ours)} & \textbf{74.32{\scriptsize $\pm$0.92}} & \textbf{49.02{\scriptsize $\pm$0.70}} & \textbf{88.63{\scriptsize $\pm$0.73}} & \textbf{72.91{\scriptsize $\pm$1.28}}  \\
\hline
\end{tabular}}
\end{subtable}
\end{table}

\subsection{Empirical Results}

\paragraph{AdaptOD vs. Other OOD Detection Methods in LTR. } 
Table \ref{fine-table} presents the comparison of our AdaptOD with two SOTA OOD detectors in LTR (EnergyOE \cite{liu2020energy}, COCL \cite{miao2023out}) on CIFAR10/100-LT using six OOD test datasets. These fine-grained results are not available for the other competing methods and thus they are omitted in this table. 
AdaptOD shows the best performance in all four metrics on each of the six OOD datasets.
Table \ref{CIFAR-OOD} shows the comparison of our AdaptOD with SOTA OOD detectors in LTR on CIFAR10/100-LT, which is the average performance over six OOD test datasets. 
Following the previous methods \cite{wang2022partial, choi2023balanced, wei2023eat}, we report our accuracy with AdjLogit \cite{menon2020long} for a fair comparison. AdaptOD is also the best performers in the averaged results when comparing to all seven competing methods.
This consistent improvement and SOTA performance of AdaptOD on both ID and OOD data indicate that the distribution gap between the outlier samples and the true OOD samples is effectively reduced by AdaptOD.
Notably, the improvement is large on the near OOD dataset CIFAR \cite{krizhevsky2009learning}, which cannot be achieved by previous SOTA methods \cite{miao2023out, choi2023balanced}.

\paragraph{DODA as an Enabler to Existing Methods. }
Table \ref{CIFAR-OOD} also presents the results of our proposed component DODA in using as a plug-in to tackle the distribution shift problem in four SOTA methods (OE, EnergyOE, BERL, and COCL) on CIFAR10/100-LT. 
It shows that DODA can consistently enhance the OOD detectors in all four metrics, demonstrating the strong capability of DODA in reducing the learned outlier distribution gap to the distribution of the true OOD data (see Appendix \ref{Experience} for more details).
The consistent improvement of having DODA as `plug-and-play' indicates the presence of the distribution shift problem encountered by existing SOTA detectors and the universal effectiveness of DODA in tackling the problem.
Note that AdaptOD as a whole achieves consistent and substantial improvement over the four DODA-enabled models, showcasing that the other component of AdaptOD, DNE, helps to learn balanced ID prediction energy and better align the adapted outlier distribution to the true OOD one.

\paragraph{AdaptOD vs. Other TTA Methods for OOD Detection. } 
Table \ref{CIFAR-TTA} shows the comparison of AdaptOD with two SOTA TTA methods AUTO and AdaODD for OOD detection on CIFAR10/100-LT. To have a straightforward and extensive comparison, we compare DODA with the two TTA methods, all of which are added on top of the same training method. In the experiments, we use five training methods, including four SOTA long-tailed OOD detection methods and our proposed DNE method. It is impressive that our DODA component consistently remains the best performer when the TTA methods are combined with all five different training methods on both ID datasets. 
DODA achieves better performance in all four OOD detection metrics across five OOD training methods, indicating that DODA is a stronger and more generic TTA method for different OOD detectors.
Moreover, the combination of our training method DNE and TTA method DODA, which is our approach AdaptOD as a whole, achieves the best performance across all 20 possible combinations.

\paragraph{Performance on Large-scale ID Data. }
To demonstrate the scalability of our approach, we also perform experiments on the large-scale ID dataset ImageNet-LT. The empirical results are presented in Table \ref{ImageNet}, which shows that our approach AdaptOD also achieves the SOTA performance in both the OOD detection performance and the ID classification accuracy.

\begin{table*}[t!]
\centering
\large
\caption{Ablation study results on CIFAR10-LT, CIFAR100-LT and ImageNet-LT.} 
\scalebox{0.58}{
\begin{tabular}{ccc|cccc|cccc|cccc}
\hline
\multirow{2}{*}{DODA} & \multirow{2}{*}{DNE-C} & \multirow{2}{*}{DNE-S} & \multicolumn{4}{c|}{ID Dataset: CIFAR10-LT} & \multicolumn{4}{c|}{ID Dataset: CIFAR100-LT} & \multicolumn{4}{c}{ID Dataset: ImageNet-LT} \\ \cline{4-15}
 &  &  & AUC$\uparrow$ & AP-in$\uparrow$ & AP-out$\uparrow$ & FPR$\downarrow$ & AUC$\uparrow$ & AP-in$\uparrow$ & AP-out$\uparrow$ & FPR$\downarrow$  & AUC$\uparrow$ & AP-in$\uparrow$ & AP-out$\uparrow$ & FPR$\downarrow$ \\
\hline
\multicolumn{3}{c|}{Baseline (EnergyOE \cite{liu2020energy})} & 91.92 & 91.03 & 91.97 & 33.80 & 76.40 & 77.32 & 72.24 & 76.33 & 69.43 & 45.12 & 84.75 & 76.89   \\
\hline
\ding{55} & \ding{55} & \ding{55} & 80.33 & 81.46 & 77.02 & 78.71 & 67.42 & 68.29 & 63.86 & 85.44 & 58.33 & 38.40 & 77.61 & 89.73   \\
\ding{51} & \ding{55} & \ding{55} & 92.63 & 92.05 & 92.46 & 30.17 & 78.10 & 80.22 & 74.17 & 71.65 & 71.71 & 45.99 & 86.37 & 74.31    \\
\ding{55} & \ding{51} & \ding{55} & 92.12  & 91.54 & 92.33 & 31.85 & 76.89 & 77.94 & 72.76 & 74.97 & 71.11 & 45.59 & 85.77 & 76.83 \\
\ding{55} & \ding{55} & \ding{51} & 91.98 & 91.36 & 91.92 & 32.44 & 76.53 & 77.46 & 72.55 & 74.62 & 70.55 & 45.36 & 84.95 & 77.02  \\
\ding{55} & \ding{51} & \ding{51} & 92.77 & 92.18 & 92.62 & 31.48 & 77.92 & 78.97 & 73.92 & 74.44 & 72.04 & 46.53 & 86.06 & 75.82  \\
\ding{51} & \ding{51} & \ding{55} & 93.81 & 93.32 & 93.53 & 28.69 & 80.07 & 82.13 & 75.73 & 68.64 & 73.14 & 47.61 & 87.19 & 73.67  \\
\ding{51} & \ding{55} & \ding{51} & 93.49 & 92.98 & 93.02 & 29.52 & 79.76 & 81.89 & 75.31 & 69.19 & 72.76 & 47.32 & 86.83 & 74.48  \\ 
\ding{51} & \ding{51} & \ding{51} & \textbf{94.69} & \textbf{93.89} & \textbf{94.12} & \textbf{27.26} & \textbf{81.93} & \textbf{83.09} & \textbf{77.83} & \textbf{67.37} & \textbf{74.32} & \textbf{49.02} & \textbf{88.63} & \textbf{72.91}\\
\hline
\multicolumn{3}{c|}{\textbf{Oracle Model}} & 95.33 & 94.75 & 94.96 & 25.02 & 83.60 & 85.09 & 78.85 & 65.37 & 75.84 & 50.20 & 89.97 & 70.71\\
\hline
\end{tabular}}
\label{ablation_table}
\end{table*}

\begin{wrapfigure}{rh!}{0.45\textwidth}
\centering
\includegraphics[width=0.4\textwidth]{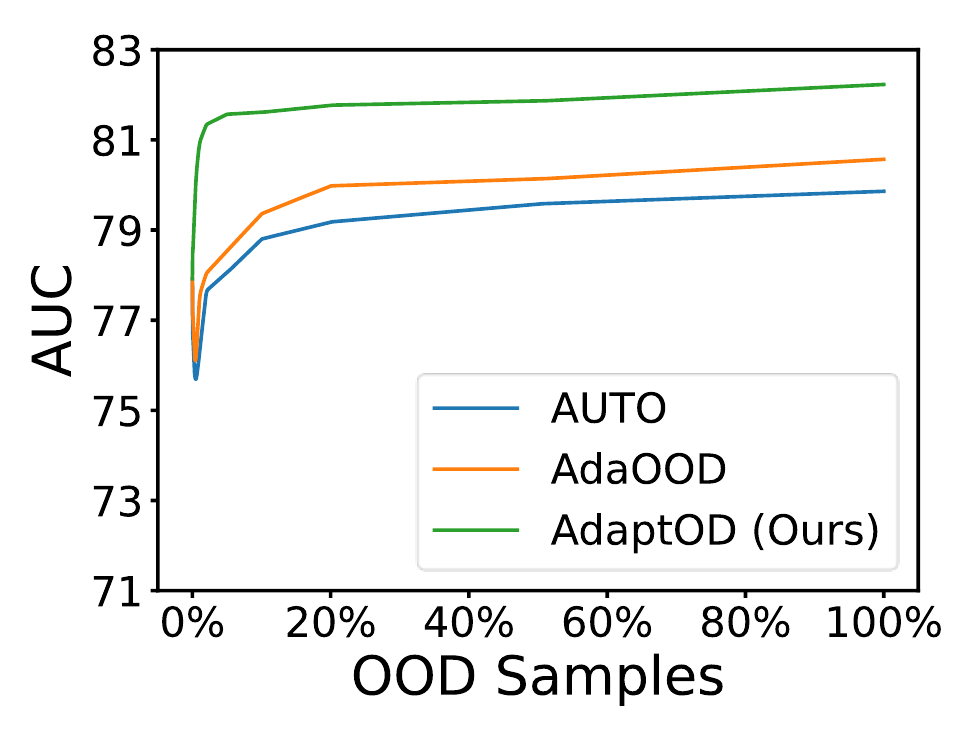} 
\caption{The average performance over six OOD datasets on CIFAR100-LT with an increasing percentage of true OOD samples fed to TTA methods.}
\label{fig3}
\vspace{-0.5cm}
\end{wrapfigure}

\subsection{Further Analysis of AdaptOD}
\paragraph{Ablation Study.}
The effectiveness of our two proposed components, DODA and DNE, have been justified in Table \ref{CIFAR-TTA}. 
Here we provide a more fine-grained analysis of DODA and its combination to two improved energy losses used in DNE, $\mathcal{L}_{C}$ (Eq.~\ref{CW-loss}, denoted as DNE-C) and $\mathcal{L}_{S}$ (Eq.~\ref{SW-loss}, denoted as DNE-S), in Table \ref{ablation_table}, with EnergyOE \cite{liu2020energy} used as baseline. 
The results show the important contribution of each component to the overall superior performance of the full model AdaptOD.
Further, we compare AdaptOD to an oracle model that utilizes the ground true OOD data to update the outlier distribution $\mathcal{P}^{out}$ in DODA. It shows that AdaptOD has only a small performance gap to the oracle model, indicating that AdaptOD can well approximate the true OOD distribution by the predicted labels of the OOD samples, without involving any ground truth during TTA.

\paragraph{OOD Data Exploitation in TTA. } 
To independently evaluate the effectiveness of exploiting OOD data to adapt the outlier distribution, we report the performance of three TTA methods with an increasing number of labeled OOD samples based on our DNE in Fig. \ref{fig3}.
All three TTA methods achieve increasing performance for OOD detection in LTR with more and more true OOD data used for the adaptation.
However, AUTO and AdaOOD struggle with the difference between training and testing ID data at the early stage of inference, while AdaptOD can utilize the adapted outlier distribution to quickly adapt to the true OOD distribution and achieve significantly improved performance.

\section{Conclusion}
To address the distribution shift problem in long-tailed OOD detection, we propose a novel approach called AdaptOD.
It utilizes a novel normalized energy-based loss -- dual-normalized energy loss (DNE) -- to learn balanced prediction energy on imbalanced ID samples and enhanced vanilla outlier distribution, then uses a dynamic outlier distribution adaptation (DODA) to adapt the outlier distribution to the true OOD distribution. DODA is shown to be a significantly improved TTA method than existing TTA methods for OOD detection. We also show that DNE can be used to support DODA with its specially designed energy training for better test-time distribution adaptation.
Experiments on three popular benchmarks demonstrated that AdaptOD significantly enhances the performance of both OOD detection and long-tailed classification.

\section*{Acknowledgments}
The participation of W. Miao, J. Zheng, and X. Bai in this work was supported by National Natural Science Foundation of China (No. 62372029 and No. 62276016), while the participation of G. Pang was supported in part by Lee Kong Chian Fellowship.

%
%
\bibliographystyle{splncs04}
\bibliography{egbib}


\newpage
\appendix

\section{More Experiment Settings}

\subsection{Datasets}
\label{datasets}
For ID datasets, the original version of CIFAR10 \cite{he2009learning} and CIFAR100 \cite{he2009learning} contains $50,000$ training images and $10,000$ validation images of size $32 \times 32$ with $10$ and $100$ classes, respectively. 
CIFAR10-LT and CIFAR100-LT are the imbalanced version of them, which reduce the number of training examples per class and keep the validation set unchanged. 
The imbalance ratio $\rho$ denotes the ratio between sample sizes of the most frequent class and least frequent class.
Following \cite{cao2019learning}, we utilize an exponential decay in sample sizes across different classes.

ImageNet-LT \cite{wu2019large} is a large-scale dataset in long-tail recognition, which truncates the balanced version ImageNet \cite{deng2009imagenet}. ImageNet-LT has $1,000$ classes, which contain $115,846$ training images with the number of per-class training data ranging from $5$ to $1,280$, and $20,000$ validation images with a balanced class size.

For outlier data, TinyImages80M \cite{torralba200880} contains $80$ million images with a size of $32×32$. We use a subset of random $30$K images as the outlier data for CIFAR10-LT and CIFAR100-LT \cite{wang2022partial,choi2023balanced}.
We use ImageNet-Extra \cite{wang2022partial} that contains $517,711$ images belonging to $500$ classes from ImageNet-22k \cite{deng2009imagenet} but having not overlapping with the $1,000$ in-distribution classes in ImageNet-LT \cite{wang2022partial,choi2023balanced}.

For OOD datasets, we use SC-OOD benchmark \cite{yang2021semantically} as true OOD data for CIFAR10-LT and CIFAR100-LT \cite{wang2022partial,choi2023balanced} following \cite{wang2022partial,choi2023balanced}. The SC-OOD benchmark contains six datasets: CIFAR \cite{krizhevsky2009learning} with $10,000$ images, Texture \cite{cimpoi2014describing} with $5,640$ images, SVHN \cite{netzer2011reading} with $26,032$ images, LSUN \cite{yu2015lsun} with $9,998$ images for CIFAR10-LT and $7,571$ images for CIFAR100-LT, Places365 \cite{zhou2017places} with $35,195$ images for CIFAR10-LT and $33,773$ images for CIFAR100-LT,  and TinyImageNet \cite{le2015tiny} with $8,793$ images for CIFAR10-LT and $7,498$ images for CIFAR100-LT. Following \cite{wang2022partial, miao2023out}, we use ImageNet-1k-OOD \cite{wang2022partial} that contains $50,000$ images belonging to $1, 000$ classes evenly from ImageNet-22k, which have not overlapping with the $1,000$ ID classes in ImageNet-LT and the $500$ outlier classes in ImageNet-Extra. A summary of the ID and OOD datasets is presented in Table \ref{data_stats}.

\begin{table*}[h]
\centering
\caption{Key statistics of the ID and OOD datasets used.} 
\scalebox{0.7}{
\begin{tabular}{c|ccc|ccc|ccc}
\hline
\multirow{2}{*}{Benchmark} & \multicolumn{3}{c|}{CIFAR10-LT} & \multicolumn{3}{c|}{CIFAR100-LT} & \multicolumn{3}{c}{ImageNet-LT} \\
 & Dataset & Images & Class & Dataset & Images & Class & Dataset & Images & Class \\
\hline
ID data (Training) & CIFAR10-LT & / & 10 & CIFAR100-LT & / & 100 & ImageNet-LT & 115,846 & 1,000 \\
ID data (Testing) & CIFAR10-LT & 10,000 & 10 & CIFAR100-LT & 10,000 & 100 & ImageNet-LT & 20,000 & 1,000 \\
\hline
Outlier data & TinyImages80M & 30,000 & / & TinyImages80M & 30,000 & / & ImageNet-Extra & 517,711 & 500 \\
\hline
\multirow{6}{*}{OOD data} & CIFAR100 & 10,000 & 100 & CIFAR10 & 10,000 & 10 & \multirow{6}{*}{\shortstack{ImageNet- \\ 1k-OOD}} & \multirow{6}{*}{50,000} & \multirow{6}{*}{1000} \\
 & Texture & 5,640 & 47 & Texture & 5,640 & 47 &  \\
 & SVHN & 26,032 & 10 & SVHN & 26,032 & 10 &  \\
 & LSUN & 9,998 & 10 & LSUN & 7,571 & / &  \\
 & Places365 & 35,195 & / & Places365 & 33,773 & / &  \\
 & TinyImageNet & 8,793 & / & TinyImageNet & 7,498 & / &  \\
\hline
\end{tabular}}
\label{data_stats}
\end{table*}

\subsection{Implementation Details}
\label{Implement}
For experiments on CIFAR10-LT \cite{cao2019learning} and CIFAR100-LT \cite{cao2019learning}, we pre-train our model based on ResNet18 \cite{he2016deep} for $320$ epochs with an initial learning rate $0.01$ \cite{alshammari2022long, choi2023balanced} using only cross-entropy loss and fine-tune the linear classifier of this model for $20$ epochs with an initial learning rate $0.001$ \cite{choi2023balanced, liu2020energy}.
The batch size is $64$ for ID data at the pre-training stage, $128$ for ID data at the fine-tuning stage, and $256$ for outlier data at the fine-tuning stage \cite{choi2023balanced, miao2023out, wang2022partial}.
Our outlier dataset is a subset of TinyImages80M \cite{torralba200880} with 30K images \cite{wang2022partial,choi2023balanced}. 

For large-scale dataset ImageNet-LT, which contains 115,846 images of 1,000 classes, we train our model based on ResNet50 \cite{he2016deep} for $100$ epochs with an initial learning rate of $0.1$ \cite{miao2023out} using only cross-entropy loss and also fine-tune the linear classifier of this model for $20$ epochs with an initial learning rate $0.01$.
Our auxiliary dataset is a subset of ImageNet22k \cite{ridnik2021imagenet}  with 516K images, following \cite{wang2022partial,miao2023out}. 

All experiments use SGD optimizer and decay the learning rate to zero using a cosine annealing learning rate scheduler \cite{loshchilov2016sgdr}. All experiences are performed with 8 NVIDIA RTX 3090.

\section{The AdaptOD Algorithm}

The full steps of the training and inference in AdaptOD are given in Algorithm \ref{alg:algorithm} below.

\label{Algo}
\begin{algorithm}[ht]
\caption{: AdaptOD}
\label{alg:algorithm}
\textbf{Training} \\
\quad \textbf{Input}:  Pre-trained model $f$ \\
\textbf{Data}:  Training dataset $ \mathcal {D}_{in}^{train} $, Auxiliary dataset $ \mathcal {D}_{out}^{train} $
\begin{algorithmic}[1] 
\FOR{each iteration} 
\STATE Sample a mini-batch of ID training data: $ \left\{ (x_i^{in}, y_i) \right\}_{i=1}^n $ from $ \mathcal {D}_{in}^{train} $
\STATE Sample a mini-batch of OOD auxiliary data: $ \left\{ (x_i^{out}) \right\}_{i=1}^n $ from $ \mathcal {D}_{out}^{train} $
\STATE Perform batch energy normalization based on Eq. \ref{Normalize}
\STATE Perform gradient descent on model $f$ with $\mathcal{L}_{total}$ based on Eq. \ref{total}
\ENDFOR
\end{algorithmic} 
\hrulefill \\
\textbf{Inference} \\
\textbf{Input}: Outlier distribution $\mathcal{P}^{out}$; Fine-tuned model $f$ \\
\textbf{Data}: Test dataset $ \mathcal {D}_{in \cup out}^{test} $ 
\begin{algorithmic}[1] 
\FOR{each sample $x$ in dataset $ \mathcal {D}_{in \cup out}^{test} $} 
\STATE Adapt the outlier distribution $\mathcal{P}^{out}$ with sample $x$ and model $f$ based on Eq. \ref{update}
\STATE Obtain calibrated global energy score $\mathbb G^{\mathcal{P}}(x)$ for sample $x$ as OOD score using the outlier distribution $\mathcal{P}^{out}$ based on Eq. \ref{prior}
\ENDFOR
\end{algorithmic}
\end{algorithm}

\section{Discussion of Stable Margin Hyperparameters in DNE}
\label{margin}
\subsection{Margin Hyperparameters in DNE-C}
To identify OOD samples, we expect ID samples to have high energy, while OOD samples to have low energy.
Formally, let $\mathbf{x}^{in} = \{ x_1^{in}, x_2^{in},..., x_{b^{in}}^{in} \}, \mathbf{x}^{in} \in X^{in}$ be one training batch of ID data, with $b^{in}$ be its batch size, and a corresponding batch of outlier data $ \mathbf{x}^{out} = \{ x_1^{out}, x_2^{out},..., x_{b^{out}}^{out} \}$ whose batch size is $b^{out}$, the sum of class-normalized energy $F_j(x)$ for all training samples in one batch (ID samples and outlier samples in one batch) in each class $j$ would be one:
\begin{equation}
\mathbf {C}_j (\mathbf{x} ^{in}) + \mathbf {C}_j (\mathbf{x} ^{out}) = \sum_{i=1}^{b^{in}} F_j(x_i^{in}) + \sum_{i=1}^{b^{out}} F_j(x_i^{out}) = 1. \label{C-vertify}
\end{equation}
DNE-C class-wisely constrains the energy that optimizes the class energy of ID samples to be large for each class, while the class energy of outlier samples is small for each class.
Therefore, the expected class-wise normalized energy for the batch of outlier samples $\mathbf {C}_j (\mathbf{x^{in}}), j \in \{1,2,...,k\}$ would be $1$ on all classes, while the expected class-wise normalized energy for the batch of ID samples $\mathbf {C}_j (\mathbf{x^{out}}), j \in \{1,2,...,k\}$ would be zero on all classes:
\begin{equation}
\begin{cases}
\mathbf {C}_j (\mathbf{x} ^{in}) \to 1, \\
\mathbf {C}_j (\mathbf{x} ^{out}) \to 0.
\label{C-proximity}
\end{cases}
\end{equation}
To this end, we set $m_{in}^c = 1$ and $m_{out}^c = 0$ for each class margin in Eq. \ref{CW-loss}, which optimizes the class-wise normalized energy for the batch of outlier samples $\mathbf {C}_j (\mathbf{x^{in}})$ on each class $j$ towards one, while at the same time optimizing the class-wise normalized energy for the batch of ID samples $\mathbf {C}_j (\mathbf{x^{out}})$ on each class $j$ towards zero.
In this way, these energy margin hyperparameters do not rely on the training dataset and/or the imbalance factor.

\subsection{Margin Hyperparameters in DNE-S}
Similarly, the sum of class-normalized energy $F_j(x)$ for all training samples in one batch (ID samples and outlier samples in one batch) in each class $j$ would be one.
Furthermore, the sum of class-normalized energy $F_j(x)$ for all training samples in one batch over all classes would be $k$ since there are $k$ categories in the ID data:
\begin{equation}
\sum_{i=1}^{b^{in}} \mathbf {S} (x_i^{in}) + \sum_{i=1}^{b^{out}} \mathbf {S} (x_i^{out}) = \sum_{i=1}^{b^{in}} \sum_{j=1}^{k} F_j(x_i^{in}) +\sum_{i=1}^{b^{out}} \sum_{j=1}^{k} F_j(x_i^{out}) = k. \label{S-vertify}
\end{equation}
DNE-S sample-wisely constrains the energy that optimizes the global energy of each ID sample to be large, while being well-balanced between head samples and tail samples. 
Therefore, the expected sample-wise normalized energy $\mathbf {S} (x), x \in \mathbf{x}^{in}$ would be $\frac{k}{b^{in}}$ for each ID sample, which is evenly divided the same scale to each ID sample (for either head samples or tail samples).
And the expected sample-wise normalized energy $\mathbf {S} (x), x \in \mathbf{x}^{out}$ would be zero for each outlier sample:
\begin{equation}
\begin{cases}
\mathbf {S} (x) \to \frac{k}{b^{in}}, x \in \mathbf{x} ^{in}, \\
\mathbf {S} (x) \to 0, x \in \mathbf{x} ^{out}.
\label{S-proximity}
\end{cases}
\end{equation}
Therefore, we set $m_{in}^s = \frac{k}{b^{in}}$ and $m_{out}^s = 0$, which optimizes the sum of class-normalized energy on all classes $\mathbf {S} (x)$ for each ID sample to $\frac{k}{b^{in}}$ and optimizes the sum of class-normalized energy on all classes $\mathbf {S} (x)$ for each outlier sample to zero. 
After doing this, we can regularize the global energy of the ID data, particularly the low global energy for tail samples, reducing the over-confident prediction of head samples. 
The same as the DNE-C loss, these specified margin parameters for training in DNE-S also do not rely on the training dataset and/or the imbalance factor.

\section{More Experimental Results}
\label{Experience}
\subsection{More Results for DODA}
Table \ref{detail-table} presents the results of our proposed component DODA in enabling two popular baselines OE \cite{hendrycks2018deep} and EnergyOE \cite{liu2020energy} on CIFAR10/100-LT on the six OOD test datasets. It shows that DODA can consistently enhance the OOD detection for both baselines in all four metrics across all six datasets, demonstrating the strong capability of DODA in reducing the learned outlier distribution gap to the true OOD. Nevertheless, these DODA-enabled baselines underperform AdaptOD, indicating that the other component of AdaptOD (\ie, DNE) helps to produce a largely enhanced vanilla outlier distribution for DODA.

\begin{table}[t]
\centering
\caption{Results of original and DODA-enabled OE and EnergyOE, and AdaptOD.}
\label{detail-table}
\scalebox{0.75}{
\begin{tabular}{c|c|cccc|cccc}
\hline
\multirow{2}{*}{\shortstack{OOD \\ Dataset}} & \multirow{2}{*}{Method} & \multicolumn{4}{c|}{ID Dataset: CIFAR10-LT} & \multicolumn{4}{c}{ID Dataset: CIFAR100-LT} \\ \cline{3-10}
 & & AUC$\uparrow$ & AP-in$\uparrow$ & AP-out$\uparrow$ & FPR$\downarrow$ & AUC$\uparrow$ & AP-in$\uparrow$ & AP-out$\uparrow$ & FPR$\downarrow$ \\ 
\hline
\multirow{5}{*}{Texture\cite{cimpoi2014describing}} & OE\cite{hendrycks2018deep} & 92.30 & 96.01 & 82.57 & 48.65 & 76.01 & 85.28 & 57.47 & 87.45 \\ 
 & OE\cite{hendrycks2018deep}+\textbf{DODA(Ours)} & \underline{95.02} & \underline{96.97} & \underline{84.40} & \underline{46.99} & \underline{77.93} &  \underline{86.51} & \underline{62.48} & \underline{82.75} \\ \cline{2-10}
 & EnergyOE\cite{liu2020energy} & 95.53 & 97.42 & 92.93 & 18.44 & 79.56 & 86.03 & 70.88 & 79.45 \\ 
 & EnergyOE\cite{liu2020energy}+\textbf{DODA(Ours)} & \underline{97.32} & \underline{97.88} & \underline{93.65} & \underline{16.28} & \underline{80.85} & \underline{87.34} & \underline{75.52} & \underline{77.00} \\ \cline{2-10}
 & \textbf{AdaptOD(Ours)} & \textbf{98.22} & \textbf{98.81} & \textbf{94.91} & \textbf{11.60} & \textbf{83.88} & \textbf{89.43} & \textbf{76.47} & \textbf{58.47} \\
\hline
\multirow{5}{*}{SVHN\cite{netzer2011reading}} & OE\cite{hendrycks2018deep} & 94.86 & 91.59 & 97.00 & 29.11 & 81.82 & 73.25 & 89.10 & 80.98 \\
 & OE\cite{hendrycks2018deep}+\textbf{DODA(Ours)} & \underline{95.95} & \underline{92.01} & \underline{98.16} & \underline{25.75} & \underline{84.20} &  \underline{75.78} & \underline{91.68} & \underline{74.86} \\ \cline{2-10}
 & EnergyOE\cite{liu2020energy} & 96.63 & 92.33 & 98.46 & 14.37 & 86.19 & 81.42 & 91.74 & 34.36 \\ 
 & EnergyOE\cite{liu2020energy}+\textbf{DODA(Ours)} & \underline{97.24} & \underline{92.88} & \underline{98.73} & \underline{12.86} &  \underline{90.26} & \underline{88.13} & \underline{95.30} & \underline{21.73}\\ \cline{2-10}
 & \textbf{AdaptOD(Ours)} & \textbf{98.13} & \textbf{94.34} & \textbf{99.11} & \textbf{10.33} & \textbf{93.09} & \textbf{91.32} & \textbf{96.86} & \textbf{17.63} \\
\hline
\multirow{5}{*}{CIFAR\cite{krizhevsky2009learning}} & OE\cite{hendrycks2018deep} & 83.32 & 84.06 & 80.83 & 65.82 & 62.60 & 66.16 & 57.77 & 93.53 \\
 & OE\cite{hendrycks2018deep}+\textbf{DODA(Ours)} & \underline{85.52} & \underline{86.03} & \underline{83.15} & \underline{60.99} & \underline{66.02} & \underline{72.11} & \underline{62.03} & \underline{90.85} \\ \cline{2-10}
 & EnergyOE\cite{liu2020energy} & 84.44 & 85.74 & 84.63 & 61.73 &  61.15 & 67.12 & 56.66 & 91.42 \\ 
 & EnergyOE\cite{liu2020energy}+\textbf{DODA(Ours)} & \underline{86.71} & \underline{87.86} & \underline{87.01} & \underline{54.33} & \underline{70.42} & \underline{76.10} & \underline{68.66} & \underline{89.87}  \\ \cline{2-10}
 & \textbf{AdaptOD(Ours)} & \textbf{89.05} & \textbf{89.93} & \textbf{88.22} & \textbf{45.51} & \textbf{72.77} & \textbf{76.37} & \textbf{70.58} & \textbf{86.04} \\
\hline
\multirow{5}{*}{\shortstack{TIN~\cite{le2015tiny}}} & OE\cite{hendrycks2018deep} & 86.35 & 89.88 & 79.30 & 64.50 & 68.22 & 79.36 & 51.82 & 88.54 \\
 & OE\cite{hendrycks2018deep}+\textbf{DODA(Ours)} & \underline{88.39} & \underline{90.88} & \underline{82.70} & \underline{61.40} & \underline{70.36} & \underline{79.72} & \underline{53.44} & \underline{88.38} \\ \cline{2-10}
 & EnergyOE\cite{liu2020energy} & 88.40 & 91.65 & 84.95 & 46.23 &  70.78 & 79.40 & 55.90 & 90.74 \\ 
 & EnergyOE\cite{liu2020energy}+\textbf{DODA(Ours)} & \underline{89.93} & \underline{92.46} & \underline{86.12} & \underline{44.02} &  \underline{71.25} & \underline{79.80} & \underline{57.91} & \underline{89.42}\\ \cline{2-10}
 & \textbf{AdaptOD(Ours)} & \textbf{91.40} & \textbf{93.85} & \textbf{88.18} & \textbf{42.77} & \textbf{72.87} & \textbf{82.06} & \textbf{58.92} & \textbf{88.24} \\
\hline
\multirow{5}{*}{LSUN\cite{yu2015lsun}} & OE & 91.57 & 93.06 & 88.37 & 53.99 & 76.81 & 85.33 & 60.94 & 83.79 \\
 & OE\cite{hendrycks2018deep}+\textbf{DODA(Ours)} & \underline{93.09} & \underline{93.42} & \underline{91.30} & \underline{48.39} & \underline{77.83} & \underline{86.24} & \underline{62.10} & \underline{82.44} \\ \cline{2-10}
 & EnergyOE\cite{liu2020energy} & 94.00 & 94.78 & 93.70 & 28.42 &  81.61 & 86.57 & 69.16 &  80.57\\ 
 & EnergyOE\cite{liu2020energy}+\textbf{DODA(Ours)} & \underline{94.92} & \underline{95.77} & \underline{94.56} & \underline{26.17} & \underline{82.54} & \underline{88.12} & \underline{70.88} & \underline{77.68} \\ \cline{2-10}
 & \textbf{AdaptOD(Ours)} & \textbf{96.16} & \textbf{96.84} & \textbf{95.86} & \textbf{24.12} & \textbf{85.70} & \textbf{90.55} & \textbf{72.70} & \textbf{70.20} \\
\hline
\multirow{5}{*}{Place365\cite{zhou2017places}} & OE\cite{hendrycks2018deep} & 90.20 & 82.09 & 95.24 & 57.06 & 75.68 & 60.99 & 86.51 & 83.55 \\
 & OE\cite{hendrycks2018deep}+\textbf{DODA(Ours)} & \underline{91.74} & \underline{83.99} & \underline{96.64} & \underline{52.57} & \underline{76.39} & \underline{62.48} & \underline{87.52} & \underline{82.72} \\ \cline{2-10}
 & EnergyOE\cite{liu2020energy} & 92.51 & 84.26 & 97.14 & 33.63 &  79.12 & 63.38 & 89.09 &  81.43\\ 
 & EnergyOE\cite{liu2020energy}+\textbf{DODA(Ours)} & \underline{94.03} & \underline{86.15} & \underline{97.75} & \underline{31.23} & \underline{81.08} & \underline{65.85} & \underline{90.94} & \underline{80.09} \\ \cline{2-10}
 & \textbf{AdaptOD(Ours)} & \textbf{95.19} & \textbf{89.56} & \textbf{98.44} & \textbf{29.22} & \textbf{83.27} & \textbf{68.82} & \textbf{91.44} & \textbf{71.63} \\
\hline
\end{tabular}}
\end{table}

\subsection{Differentiating OOD Data from Head and Tail Samples.} 
To evaluate the effectiveness in distinguishing OOD data from head and tail samples, we perform two particular inference settings: one with only tail samples and OOD samples, and another one with only head samples and OOD samples. 
Table \ref{Part} shows the averaged results over the six OOD test datasets on CIFAR10/100-LT of the baseline EnergyOE \cite{liu2020energy}, previous SOTA model COCL \cite{miao2023out}, and our AdaptOD.
It can be observed that AdaptOD does a better job than the two methods in both scenarios, resulting in significantly enhanced overall detection performance.

\begin{table}[t]
\centering
\caption{\centering{Comparison results on separating head/tail samples from OOD samples.}}
\label{Part}
\scalebox{0.8}{
\begin{tabular}{c|c|cccc|cccc}
\hline
\multirow{2}{*}{ID dataset} & \multirow{2}{*}{method} & \multicolumn{4}{c|}{Head Samples} & \multicolumn{4}{c}{Tail Samples}  \\ \cline{3-10}
 &  & AUC$\uparrow$ & AP-in$\uparrow$ & AP-out$\uparrow$ & FPR$\downarrow$ & AUC$\uparrow$ & AP-in$\uparrow$ & AP-out$\uparrow$ & FPR$\downarrow$   \\
\hline
\multirow{3}{*}{CIFAR10-LT } & EnergyOE\cite{liu2020energy} & 95.88 & 89.67 &  98.31 & 23.06 & 83.45  & 61.07 & 93.37 & 58.61 \\
 & COCL\cite{miao2023out} & 96.34 & 93.34 & 98.67 & 19.59 & 91.91 & 76.98 & 97.15 & 34.30 \\
 & \textbf{AdaptOD(Ours)} & \textbf{98.20} & \textbf{96.80} & \textbf{99.00} & \textbf{11.20} & \textbf{93.40} & \textbf{80.58} & \textbf{98.27} & \textbf{30.70} \\
\hline
\multirow{3}{*}{CIFAR100-LT } & EnergyOE\cite{liu2020energy} & 84.22 & 69.70 &  92.81 & 69.42 & 67.63 & 35.85 & 85.96 & 81.77  \\
 & COCL\cite{miao2023out} & 87.73 & 73.84 & 93.94 & 66.01 & 74.85 & 47.76 & 87.59 & 77.01 \\
 & \textbf{AdaptOD(Ours)} & \textbf{91.81} & \textbf{80.42} & \textbf{96.43} & \textbf{58.49} & \textbf{78.34} & \textbf{56.67} & \textbf{91.15} & \textbf{70.82} \\
\hline
\end{tabular}}
\end{table}

\subsection{More Ablation Study}
\label{Ablation}
\subsubsection{Imbalance Ratio}
In the Experiments section, we use the default imbalance ratio $\rho=100$ on both CIFAR10-LT and CIFAR100-LT. In this section, we show that our method can work well under different imbalance ratios.
Table \ref{Imbalance_ratio} shows the comparison of AdaptOD with two SOTA long-tailed OOD detection methods EnergyOE \cite{liu2020energy} and COCL \cite{miao2023out} on CIFAR10-LT with $\rho=50$ and $\rho=10$.
Our approach can significantly outperform these baselines in not only OOD detection performance but also ID classification accuracy by a considerable margin with different imbalance ratios.
Furthermore, our approach AdaptOD performs better in more imbalanced datasets, indicating the superiority of AdaptOD for OOD detection in long-tail recognition.

Table \ref{imbalance-TTA} shows the comparison of our approach AdaptOD with two SOTA TTA methods AUTO and AdaODD for OOD detection on CIFAR10-LT with $\rho=50$. 
To have a straightforward and extensive comparison, we compare DODA (the component of AdaptOD) with the two TTA methods, all of which are added on top of the same training method. 
In the experiments, we use three training methods, including two SOTA long-tailed OOD detection methods and our proposed DNE-based training method. 
It is impressive that our DODA component consistently remains the best performer when the TTA methods are combined with all three different training methods on the CIFAR10-LT datasets. 

\begin{table}[t!]
\caption{Comparison results of imbalance ratio among EnergyOE \cite{liu2020energy}, COCL \cite{miao2023out}, and our approach AdaptOD on CIFAR10-LT. 
}
\centering
\scalebox{0.8}{
\begin{tabular}{c|c|ccccc}
\hline
Imbalance Ratio& Method & AUC$\uparrow$ & AP-in$\uparrow$ & AP-out$\uparrow$ & FPR$\downarrow$  & ACC$\uparrow$ \\ 
\hline
\multirow{3}{*}{$\rho=100$} & EnergyOE \cite{liu2020energy} & 91.92 & 91.03 & 91.97 & 33.80 & 74.57 \\
& COCL \cite{miao2023out} & 93.28 & 92.24 & 92.89 & 30.88 & 81.56 \\
& \textbf{AdaptOD(Ours)} & \textbf{94.69} & \textbf{93.89} & \textbf{94.12} & \textbf{27.26} & \textbf{82.27} \\
\hline
\multirow{3}{*}{$\rho=50$} & EnergyOE \cite{liu2020energy} & 93.48 & 92.68 & 93.05 & 29.74 & 81.23 \\
& COCL \cite{miao2023out} & 94.30 & 93.85 & 93.31 & 26.98 & 84.89 \\
& \textbf{AdaptOD(Ours)} & \textbf{95.14} & \textbf{94.53} & \textbf{94.66} & \textbf{24.43} & \textbf{85.77} \\
\hline
\multirow{3}{*}{$\rho=10$} & EnergyOE \cite{liu2020energy} & 95.03 & 94.34 & 94.83 & 25.26 & 88.47 \\
& COCL \cite{miao2023out} & 95.71 & 95.12 & 95.33 & 20.91 & 89.65\\
& \textbf{AdaptOD(Ours)} & \textbf{96.34} & \textbf{95.72} & \textbf{95.86} & \textbf{18.33} & \textbf{90.24} \\
\hline
\end{tabular}
}
\label{Imbalance_ratio}
\end{table}

\begin{table}[t!]
\centering
\caption{Comparison to different TTA-based OOD detection methods on CIFAR10-LT with $\rho = 50$. }
\label{imbalance-TTA}
\scalebox{0.8}{
\begin{tabular}{cc|cccc}
\hline
Training & Test  & AUC$\uparrow$ & AP-in$\uparrow$ & AP-out$\uparrow$ & FPR$\downarrow$ \\ 
\hline
\multirow{4}{*}{EnergyOE\cite{liu2020energy}} & w/o TTA & 93.48$\pm0.25$ & 92.68$\pm$0.33 & 93.05$\pm$0.30 & 29.74$\pm$0.22  \\
 & AUTO\cite{yang2023auto} & 93.85$\pm$0.29 & 92.84$\pm$0.25 & 93.34$\pm$0.9 & 29.10$\pm$0.35  \\
 & AdaODD\cite{zhang2023model} & 94.14$\pm$0.33 & 92.92$\pm$0.32 & 93.60$\pm$0.33 & 29.01$\pm$0.20  \\
 & \textbf{DODA(Ours)} & \underline{94.60$\pm$0.28} & \underline{93.46$\pm$0.36} & \underline{93.91$\pm$0.26} & \underline{28.42$\pm$0.24} \\
\hline
\multirow{4}{*}{COCL\cite{miao2023out}} & w/o TTA  & 94.30$\pm$0.25 & 93.85$\pm$0.25 & 93.31$\pm$0.44 & 26.98$\pm$0.28   \\
 & AUTO\cite{yang2023auto} & 94.62$\pm$0.31 & 93.91$\pm$0.33 & 93.52$\pm$0.40 & 26.49$\pm$0.37  \\
 & AdaODD\cite{zhang2023model} & 94.41$\pm$0.29 & 93.84$\pm$0.31 & 93.35$\pm$0.46 & 26.67$\pm$0.35  \\
 & \textbf{DODA(Ours)} & \underline{94.82$\pm$0.24} & \underline{94.13$\pm$0.29} & \underline{94.21$\pm$0.36} & \underline{26.02$\pm$0.30} \\
\hline
\multirow{4}{*}{\textbf{\shortstack{DNE \\ (Ours)}}}  & w/o TTA  & 93.85$\pm$0.38 & 93.43$\pm$0.28 & 93.62$\pm$0.39 & 27.47$\pm$0.38   \\
 & AUTO\cite{yang2023auto} & 94.59$\pm$0.43 & 93.68$\pm$0.33 & 93.98$\pm$0.38 & 26.50$\pm$0.35  \\
 & AdaODD\cite{zhang2023model} & 94.75$\pm$0.42 & 93.78$\pm$0.32 & 94.22$\pm$0.44 & 25.64$\pm$0.37  \\
 & \textbf{DODA(Ours)} & \textbf{95.14$\pm$0.41} & \textbf{94.53$\pm$0.27} & \textbf{94.66$\pm$0.36} & \textbf{24.43$\pm$0.32} \\
\hline
\end{tabular}
}
\end{table}

\begin{table}[t!]
\caption{Comparison results of model structure among EnergyOE \cite{liu2020energy}, COCL \cite{miao2023out}, and our approach AdaptOD on CIFAR10-LT. }
\centering
\scalebox{0.8}{
\begin{tabular}{c|c|c|c|c|c|c}
\hline
Model& Method & AUC$\uparrow$ & AP-in$\uparrow$ & AP-out$\uparrow$ & FPR$\downarrow$ & ACC$\uparrow$ \\ 
\hline
\multirow{3}{*}{ResNet18} & EnergyOE \cite{liu2020energy} & 91.92 & 91.03 & 91.97 & 33.80 & 74.57 \\
& COCL \cite{miao2023out} & 93.28 & 92.24 & 92.89 & 30.88 & 81.56\\
& \textbf{AdaptOD(Ours)} & \textbf{94.69} & \textbf{93.89} & \textbf{94.12} & \textbf{27.26} & \textbf{82.27} \\
\hline
\multirow{3}{*}{ResNet34} & EnergyOE \cite{liu2020energy} & 92.25 & 91.37 & 92.31 & 32.44 & 74.89 \\
& COCL \cite{miao2023out} & 93.52 & 92.93 & 92.83 & 30.74 & 81.75\\
& \textbf{AdaptOD(Ours)} & \textbf{94.98} & \textbf{94.33} & \textbf{94.52} & \textbf{26.61} & \textbf{83.47} \\
\hline
\end{tabular}
}
\label{Model_structures_cifar10}
\end{table}

\begin{table}[t!]
\centering
\caption{Comparison to different TTA-based OOD detection methods on CIFAR10-LT using ResNet34. The results are averaged over the six OOD test datasets in the SC-OOD benchmark.}
\label{structure-TTA}
\scalebox{0.8}{
\begin{tabular}{cc|cccc}
\hline
Training & Test  & AUC$\uparrow$ & AP-in$\uparrow$ & AP-out$\uparrow$ & FPR$\downarrow$ \\ 
\hline
\multirow{4}{*}{EnergyOE\cite{liu2020energy}} & w/o TTA & 92.25$\pm$0.32 & 91.37$\pm$0.31 & 92.31$\pm$0.28 & 32.44$\pm$0.37  \\
 & AUTO\cite{yang2023auto} & 92.98$\pm$0.40 & 91.93$\pm$0.26 & 92.60$\pm$0.38 & 32.03$\pm$0.38  \\
 & AdaODD\cite{zhang2023model} & 93.16$\pm$0.36 & 92.10$\pm$0.50 & 92.90$\pm$0.46 & 31.87$\pm$0.47  \\
 & \textbf{DODA(Ours)} & \underline{93.81$\pm$0.32} & \underline{92.68$\pm$0.28} & \underline{93.27$\pm$0.39} & \underline{29.65$\pm$0.36} \\
\hline
\multirow{4}{*}{COCL\cite{miao2023out}} & w/o TTA  & 93.52$\pm$0.36 & 92.93$\pm$0.48 & 92.83$\pm$0.27 & 30.74$\pm$0.38   \\
 & AUTO\cite{yang2023auto} & 93.73$\pm$0.45 & 93.04$\pm$0.40 & 93.26$\pm$0.31 & 29.60$\pm$0.40  \\
 & AdaODD\cite{zhang2023model} & 93.90$\pm$0.47 & 93.19$\pm$0.35 & 93.46$\pm$0.44 & 29.31$\pm$0.41  \\
 & \textbf{DODA(Ours)} & \underline{94.27$\pm$0.39} & \underline{93.57$\pm$0.38} & \underline{93.82$\pm$0.36} & \underline{28.78$\pm$0.34} \\
\hline
\multirow{4}{*}{\textbf{\shortstack{DNE \\ (Ours)}}}  & w/o TTA  & 93.28$\pm$0.30 & 92.64$\pm$0.25 & 92.95$\pm$0.32 & 31.18$\pm$0.33   \\
 & AUTO\cite{yang2023auto} & 93.77$\pm$0.33 & 92.83$\pm$0.50 & 93.36$\pm$0.45 & 29.99$\pm$0.47  \\
 & AdaODD\cite{zhang2023model} & 93.84$\pm$0.42 & 93.04$\pm$0.39 & 93.61$\pm$0.37 & 29.53$\pm$0.42  \\
 & \textbf{DODA(Ours)} & \textbf{94.98$\pm$0.35} & \textbf{94.33$\pm$0.33} & \textbf{94.52$\pm$0.40} & \textbf{26.61$\pm$0.37} \\
\hline
\end{tabular}
}
\end{table}

\subsubsection{Network Architectures}
In the Experiments section, we use the standard ResNet18 as the backbone model on both CIFAR10-LT and CIFAR100-LT. To show the generality of our method, we also perform a long-tailed OOD detection experiment using both ResNet34 and ResNet18.
The results are shown in Table \ref{Model_structures_cifar10} and Table \ref{structure-TTA}. 
Table \ref{Model_structures_cifar10} shows the comparison of AdaptOD with two SOTA long-tailed OOD detection methods EnergyOE \cite{liu2020energy} and COCL \cite{miao2023out} on CIFAR10-LT using different backbone models.
Table \ref{structure-TTA} shows the comparison of AdaptOD with two SOTA TTA methods AUTO and AdaODD for OOD detection on CIFAR10-LT using ResNet34. AdaptOD maintains its superiority with different network architectures.

\subsubsection{Sensitivity} 
\label{ablation_a}
\begin{figure}[h]
    \centering
    \includegraphics[width=0.4\columnwidth]{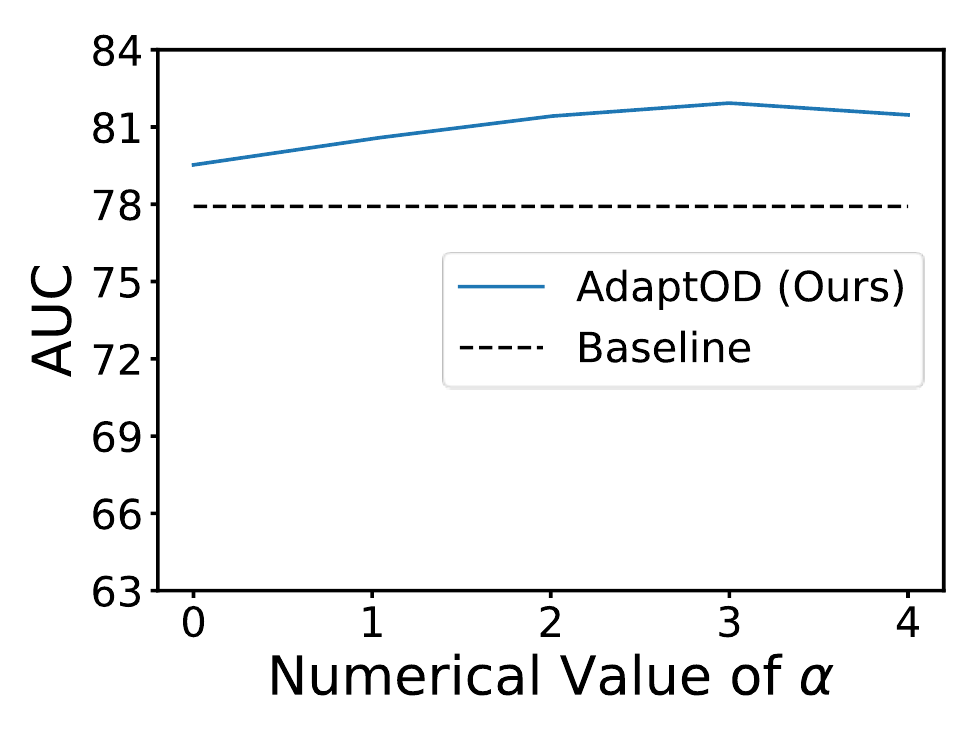}
    \caption{Average performance of AdaptOD w.r.t. $\alpha$ over six OOD datasets on CIFAR100-LT.}
    \label{K}
\end{figure}

In the Approach section, we utilize a Z-score-based method based on training ID data to implement the OOD filter, which predicts true OOD samples for adapting outlier distribution in DODA.
The threshold $R$ in the OOD filter is calculated with only training ID data and can be directly used during inference.
$\alpha$ is a hyperparameter to adjust the threshold $R$. 
A too high value of $R$ can misclassify a large number of ID samples as OOD samples.
On the other hand, a too low value of $R$ will filter out too many true OOD samples. In both cases, the outlier distribution adaptation becomes ineffective.
Fig. \ref{K} shows the sensitivity of AdaptOD with respect to $\alpha$ in Eq.~\ref{filter}, showing that the performance of AdaptOD is relatively stable with a relatively large range of $\alpha$ values, \eg, $[2.5, 3.5]$.

\subsubsection{Computational Overhead} 
AdaptOD performs normalization on the logit output of each class for each batch of training samples before energy training.
Fig. \ref{compute} shows the computational overhead of AdaptOD compared to the baseline EnergyOE and the previous SOTA method BERL using the same backbone on CIFAR100-LT with a single NVIDIA RTX 3090, in which they are all fine-tuning-based and energy-based methods.
It shows that the training speed of AdaptOD is similar with the previous methods on both ResNet18 backbone and ResNet34 backbones.
\begin{table}[h]
\caption{Comparison results of training time (seconds) on CIFAR100-LT.}
\centering
\scalebox{0.8}{
\begin{tabular}{c|c|c|c}
\hline
\multirow{2}{*}{Model} & \multicolumn{3}{c}{Training Time (seconds)} \\
\cline{2-4}
 & EnergyOE\cite{liu2020energy} & BERL \cite{choi2023balanced} & AdaptOD(Ours)\\ 
\hline
ResNet18 & 8.56 & 9.12 & 8.84 \\
\hline
ResNet34 & 12.65 & 13.12 & 12.89 \\
\hline
\end{tabular}
}
\label{compute}
\end{table}

\begin{table}[h]
\caption{Comparison results on synthetic OOD datasets with CIFAR10-LT.}
\centering
\scalebox{0.8}{
\begin{tabular}{c|c|c|c|c|c}
\hline
Dataset & Method & AUC$\uparrow$ & AP-in$\uparrow$ & AP-out$\uparrow$ & FPR$\downarrow$ \\ 
\hline
\multirow{4}{*}{Gaussian} & EnergyOE \cite{liu2020energy} & 99.74 & 99.76 & 99.33 & 1.96\\
& BERL \cite{choi2023balanced}  & 99.76 & 99.34 & 99.16 & 0.49\\ 
& COCL \cite{miao2023out} & 99.68 & 99.79 & 99.39 & \textbf{0.02} \\
& \textbf{AdaptOD(Ours)} & \textbf{99.83} & \textbf{99.87} & \textbf{99.58} & 0.08 \\
\hline
\multirow{4}{*}{Rademacher} & EnergyOE \cite{liu2020energy} & 99.13 & 99.25 & 97.16 & 2.32\\
& BERL \cite{choi2023balanced} & 99.00 & 99.06 & 96.26 & 1.42 \\ 
& COCL \cite{miao2023out} & 99.76 & \textbf{99.84} & 99.56 & \textbf{0.01} \\
& \textbf{AdaptOD(Ours)} & \textbf{99.78} & 99.65 & \textbf{99.61} & 0.04 \\
\hline
\multirow{4}{*}{Blobs} & EnergyOE \cite{liu2020energy} & 90.16 & 93.25 & 85.39 & 9.44 \\
& BERL \cite{choi2023balanced} & 93.18 & 96.87 & 89.34 & 6.54 \\ 
& COCL \cite{miao2023out} & 98.75 & 99.17 & 97.49 & 1.04 \\
& \textbf{AdaptOD(Ours)} & \textbf{99.12} & \textbf{99.43} & \textbf{98.62} & \textbf{0.53} \\
\hline
\multirow{4}{*}{Average} & EnergyOE \cite{liu2020energy} & 96.34 & 97.42 & 93.96 & 4.57\\
& BERL \cite{choi2023balanced} & 97.32 & 98.42 & 94.92 & 2.81 \\ 
& COCL \cite{miao2023out} & 99.40 & 99.60 & 98.81 & 0.35 \\
& \textbf{AdaptOD(Ours)} & \textbf{99.58} & \textbf{99.65} & \textbf{99.27} & \textbf{0.22} \\
\hline
\end{tabular}
}
\label{synthetic}
\end{table}

\subsection{Experiment Results on Synthetic OOD Datasets}
To demonstrate the superiority of our approach AdaptOD on diverse OOD datasets, we also evaluate our approach AdaptOD with three synthetic OOD datasets on CIFAR10-LT, including Gaussian, Rademacher, and Blobs. 
Specifically, \textit{Gaussian} noises have each dimension sampled from an isotropic Gaussian distribution. \textit{Rademacher} noises are images where each dimension is $\textnormal{-} 1$ or $1$ with equal probability, so each dimension is sampled from a symmetric Rademacher distribution. \textit{Blobs} noises consist of algorithmically generated amorphous shapes with definite edges. 
We use three SOTA methods for comparison, including EnergyOE \cite{liu2020energy}, BERL \cite{choi2023balanced}, and COCL \cite{miao2023out}. 
As in Table~\ref{synthetic}, our approach AdaptOD achieves similarly significant improvement over these methods on these synthetic OOD datasets as on the other OOD datasets.

\section{Limitation and Broader Impacts}
\subsection{Limitation}
\label{Limitation}
While AdaptOD offers a straightforward and competitive solution for out-of-distribution detection in long-tailed recognition, it necessitates the incorporation of additional outlier data to learn an enhanced vanilla outlier distribution, increasing the difficulty of applying it to real-world scenarios. 
The DODA component in AdaptOD is an attempt that utilizes the detected OOD samples during the inference stage to improve OOD detection performance. This requires online updating of the learned outlier distribution. An alternative way is to optimize the outlier distribution and reduce its gap to the OOD distribution during training, eliminating the online updating step. The lack of true OOD data remains as a major challenge in such approaches. We leave it for future work. 

\subsection{Broader Impacts}
\label{Impacts}
OOD detection is a branch of anomaly detection that typically plays a positive role in enhancing model security in various safety-critical applications such as autonomous driving. 
When applying our methods, we need to ensure that the models are only used for the purpose of enhancing the safety of deep learning models in real-world environments and do not infringe on human privacy.


\clearpage
\section*{NeurIPS Paper Checklist}

\begin{enumerate}

\item {\bf Claims}
    \item[] Question: Do the main claims made in the abstract and introduction accurately reflect the paper's contributions and scope?
    \item[] Answer: \answerYes{} 
    \item[] Justification: Abstract and introduction accurately reflect the paper's contributions and scope.

\item {\bf Limitations}
    \item[] Question: Does the paper discuss the limitations of the work performed by the authors?
    \item[] Answer: \answerYes{} 
    \item[] Justification: The limitations of the work are discussed in Appendix \ref{Limitation}.

\item {\bf Theory Assumptions and Proofs}
    \item[] Question: For each theoretical result, does the paper provide the full set of assumptions and a complete (and correct) proof?
    \item[] Answer: \answerNA{} 
    \item[] Justification: There is no theoretical result in this paper.

    \item {\bf Experimental Result Reproducibility}
    \item[] Question: Does the paper fully disclose all the information needed to reproduce the main experimental results of the paper to the extent that it affects the main claims and/or conclusions of the paper (regardless of whether the code and data are provided or not)?
    \item[] Answer: \answerYes{} 
    \item[] Justification: Our paper fully discloses all the information needed to reproduce the main experimental results.

\item {\bf Open access to data and code}
    \item[] Question: Does the paper provide open access to the data and code, with sufficient instructions to faithfully reproduce the main experimental results, as described in supplemental material?
    \item[] Answer: \answerYes{} 
    \item[] Justification: Our paper provides open access to the data and code.

\item {\bf Experimental Setting/Details}
    \item[] Question: Does the paper specify all the training and test details (e.g., data splits, hyperparameters, how they were chosen, type of optimizer, etc.) necessary to understand the results?
    \item[] Answer: \answerYes{} 
    \item[] Justification: Our paper specifies all the training and testing details

\item {\bf Experiment Statistical Significance}
    \item[] Question: Does the paper report error bars suitably and correctly defined or other appropriate information about the statistical significance of the experiments?
    \item[] Answer: \answerYes{} 
    \item[] Justification: Our paper shows average results over six runs with different random seeds and report the variance for our metrics.

\item {\bf Experiments Compute Resources}
    \item[] Question: For each experiment, does the paper provide sufficient information on the computer resources (type of compute workers, memory, time of execution) needed to reproduce the experiments?
    \item[] Answer: \answerYes{} 
    \item[] Justification: Our paper provides sufficient information on the computer resources.
    
\item {\bf Code Of Ethics}
    \item[] Question: Does the research conducted in the paper conform, in every respect, with the NeurIPS Code of Ethics \url{https://neurips.cc/public/EthicsGuidelines}?
    \item[] Answer: \answerYes{} 
    \item[] Justification: Our research complies with the NeurIPS Code of Ethics.

\item {\bf Broader Impacts}
    \item[] Question: Does the paper discuss both potential positive societal impacts and negative societal impacts of the work performed?
    \item[] Answer: \answerYes{} 
    \item[] Justification: The societal impacts of the work are discussed in Appendix \ref{Impacts}.
    
\item {\bf Safeguards}
    \item[] Question: Does the paper describe safeguards that have been put in place for responsible release of data or models that have a high risk for misuse (e.g., pretrained language models, image generators, or scraped datasets)?
    \item[] Answer: \answerNA{} 
    \item[] Justification: The paper poses no such risk.

\item {\bf Licenses for existing assets}
    \item[] Question: Are the creators or original owners of assets (e.g., code, data, models), used in the paper, properly credited and are the license and terms of use explicitly mentioned and properly respected?
    \item[] Answer: \answerYes{} 
    \item[] Justification: Our paper gives proper acknowledgement of the original papers that produced the code packages or datasets.

\item {\bf New Assets}
    \item[] Question: Are new assets introduced in the paper well documented and is the documentation provided alongside the assets?
    \item[] Answer: \answerNA{} 
    \item[] Justification: Our paper does not release new assets.

\item {\bf Crowdsourcing and Research with Human Subjects}
    \item[] Question: For crowdsourcing experiments and research with human subjects, does the paper include the full text of instructions given to participants and screenshots, if applicable, as well as details about compensation (if any)? 
    \item[] Answer: \answerNA{} 
    \item[] Justification: Our paper does not involve crowdsourcing nor research with human subjects.

\item {\bf Institutional Review Board (IRB) Approvals or Equivalent for Research with Human Subjects}
    \item[] Question: Does the paper describe potential risks incurred by study participants, whether such risks were disclosed to the subjects, and whether Institutional Review Board (IRB) approvals (or an equivalent approval/review based on the requirements of your country or institution) were obtained?
    \item[] Answer: \answerNA{} 
    \item[] Justification: Our paper does not involve crowdsourcing nor research with human subjects.

\end{enumerate}

\end{sloppypar}

\end{document}